\documentclass{article}

\usepackage{microtype}
\usepackage{graphicx}
\usepackage{subfigure}
\usepackage{booktabs} % for professional tables
\usepackage{array}
\usepackage{makecell} 
\usepackage{multirow}
\usepackage{hyperref}
\usepackage{algorithmicx}
\usepackage{algpseudocode}
\usepackage{xcolor}
\usepackage{wrapfig}
\usepackage{tikz}
\definecolor{myblue}{RGB}{69, 148, 193} % 定义自定义颜色

\usepackage[accepted]{icml2025}

\usepackage{amsmath}
\usepackage{amssymb}
\usepackage{mathtools}
\usepackage{amsthm}

\usepackage[capitalize,noabbrev]{cleveref}

\theoremstyle{plain}

\theoremstyle{definition}

\theoremstyle{remark}

\usepackage[textsize=tiny]{todonotes}

\icmltitlerunning{Neuron-level Balance between Stability and Plasticity in Deep Reinforcement Learning}

\begin{document}

\twocolumn[
\icmltitle{Neuron-level Balance between Stability and Plasticity \\ in Deep Reinforcement Learning}

% It is OKAY to include author information, even for blind
% submissions: the style file will automatically remove it for you
% unless you've provided the [accepted] option to the icml2025
% package.

% List of affiliations: The first argument should be a (short)
% identifier you will use later to specify author affiliations
% Academic affiliations should list Department, University, City, Region, Country
% Industry affiliations should list Company, City, Region, Country

% You can specify symbols, otherwise they are numbered in order.
% Ideally, you should not use this facility. Affiliations will be numbered
% in order of appearance and this is the preferred way.
\icmlsetsymbol{equal}{*}

\begin{icmlauthorlist}
\icmlauthor{Jiahua Lan}{yyy}
\icmlauthor{Sen Zhang}{comp}
\icmlauthor{Haixia Pan}{yyy}
\icmlauthor{Ruijun Liu}{yyy}
\icmlauthor{Li Shen}{sch}
\icmlauthor{Dacheng Tao}{ntu}
% \icmlauthor{Firstname7 Lastname7}{comp}
%\icmlauthor{}{sch}
% \icmlauthor{Firstname8 Lastname8}{sch}
% \icmlauthor{Firstname8 Lastname8}{yyy,comp}
%\icmlauthor{}{sch}
%\icmlauthor{}{sch}
\end{icmlauthorlist}

\icmlaffiliation{yyy}{School of Software, Beihang University, Beijing, China}
\icmlaffiliation{comp}{TikTok, Sydney, Australia}
\icmlaffiliation{sch}{School of Cyber Science and Technology, Shenzhen Campus of Sun Yat-sen University, Shenzhen, China}
\icmlaffiliation{ntu}{Nanyang Technological University, Singapore}

\icmlcorrespondingauthor{Haixia Pan}{haixiapan@buaa.edu.cn}

% You may provide any keywords that you
% find helpful for describing your paper; these are used to populate
% the "keywords" metadata in the PDF but will not be shown in the document
% \icmlkeywords{Machine Learning, ICML}

\vskip 0.3in
]

% this must go after the closing bracket ] following \twocolumn[ ...

% This command actually creates the footnote in the first column
% listing the affiliations and the copyright notice.
% The command takes one argument, which is text to display at the start of the footnote.
% The \icmlEqualContribution command is standard text for equal contribution.
% Remove it (just {}) if you do not need this facility.

\printAffiliationsAndNotice{}  % leave blank if no need to mention equal contribution
% \printAffiliationsAndNotice{\icmlEqualContribution} % otherwise use the standard text.

\begin{abstract}

In contrast to the human ability to continuously acquire knowledge, agents struggle with the stability-plasticity dilemma in deep reinforcement learning (DRL), which refers to the trade-off between retaining existing skills (stability) and learning new knowledge (plasticity).
Current methods focus on balancing these two aspects at the network level, lacking sufficient differentiation and fine-grained control of individual neurons.
To overcome this limitation, we propose Neuron-level Balance between Stability and Plasticity (NBSP) method, by taking inspiration from the observation that specific neurons are strongly relevant to task-relevant skills. 
%To the best of our knowledge, this is the first work to tackle both stability and plasticity loss in DRL at the level of neurons. 
Specifically, NBSP first (1) defines and identifies RL skill neurons that are crucial for knowledge retention through a goal-oriented method, and then (2) introduces a framework by employing gradient masking and experience replay techniques targeting these neurons to preserve the encoded existing skills while enabling adaptation to new tasks.
Numerous experimental results on the Meta-World and Atari benchmarks demonstrate that NBSP significantly outperforms existing approaches in balancing stability and plasticity. 
% Furthermore, our findings underscore the pivotal role of the critic within this context, providing valuable insights for future research.
\end{abstract}

\section{Introduction}
\label{introduction}

Deep reinforcement learning (DRL) has shown exceptional capabilities across a range of complex scenarios, such as gaming \citep{atari}, robotic manipulation \citep{robot}, and autonomous driving \citep{autodriveing}. However, most RL research focuses on agents that learn to solve individual problems rather than agents that learn continually. When agent try to learn a sequence of tasks continually, the \textbf{stability-plasticity dilemma} remains a fundamental and under-explored problem. Ideally, the agent must maintain its performance on previously learned tasks, a characteristic referred to as \textbf{stability} \citep{stability}, while simultaneously adapting to new tasks, known as \textbf{plasticity} \citep{plasticity}. However, it has been revealed that emphasizing stability may hinder the ability of agents to learn new knowledge \citep{plasticity1, plasticity2}, whereas excessive plasticity can lead to catastrophic forgetting of previously acquired knowledge \citep{stability1, stability2}, a challenge known as the \textbf{stability-plasticity dilemma} \citep{dilemma}, which is the main focus of our work.

% , which remains under-explored in the research community. CRL holds significant potential for developing incremental learners capable of adapting to dynamic and realistic applications, including education, logistics, and robotics.  While naive solutions such as retraining an agent from scratch and developing a new agent for each task are either computationally expensive or impractical, a more efficient alternative is to enable agents to learn new tasks continually based on the knowledge acquired from the previous ones. 

% Despite the promise of the continual RL approach, the \textbf{stability-plasticity dilemma} remains a fundamental and under-explored problem. Ideally, the agent must maintain its performance on previously learned tasks, a characteristic referred to as \textbf{stability}\citep{stability}, while simultaneously adapting to new tasks, known as \textbf{plasticity} \citep{plasticity}. However, it has been revealed that emphasizing stability may hinder the ability of agents to learn new knowledge \citep{plasticity1, plasticity2}, whereas excessive plasticity can lead to catastrophic forgetting of previously acquired knowledge \citep{stability1, stability2}, a challenge known as the \textbf{stability-plasticity dilemma} \citep{dilemma}, which is the main focus of our work.

Existing methods to strike a balance between stability and plasticity  generally fall into three categories, i.e. 
(1) \textbf{regularization-based methods} \citep{ewc, regularization}, which apply penalties to parameter changes to mitigate forgetting while acquiring new knowledge;  
(2) \textbf{replay-based methods} \citep{replay}, which leverage past experiences to consolidate knowledge; and  
(3) \textbf{modularity-based methods} \citep{aux, modularity1}, which seek to decouple stability and plasticity or isolate different components for different tasks. 
Despite their contributions, these methods suffer from three key limitations: (1) They primarily operate at the network level, yet their ultimate impact manifests at the level of individual neurons. However, these methods fail to differentiate and fine-grained control neurons based on their specific roles. Therefore, identifying and effectively utilizing task-relevant neurons remains both critical and under-explored. (2) These studies are primarily conducted within the framework of continual learning, thus overlooking the unique characteristics intrinsic to DRL. (3) These approaches could sometimes unnecessarily inflate model parameters, thereby introducing unwarranted complexity \citep{modularity2}.

\begin{figure}[t]
% \centering
\begin{center}
%\framebox[4.0in]{$\;$}
\vspace{0.5em}
\includegraphics[width=0.9\columnwidth]{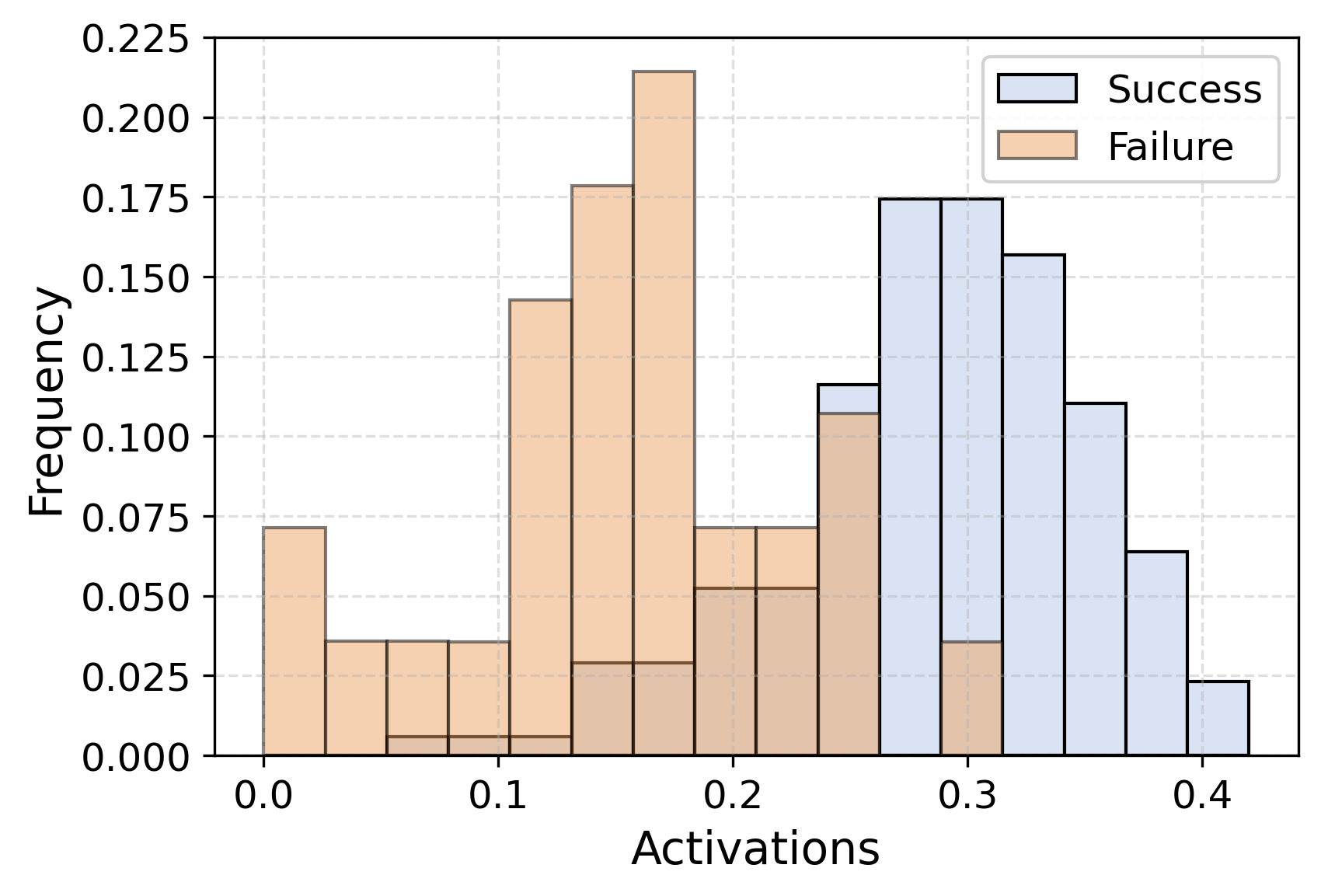}
\vspace{-1.5em}
\end{center}
\caption{Distribution histogram of the activation of a neuron, categorized based on whether the drawer-open task was successfully completed or not.}
\label{fig:activation}
\vspace{-1em}
\end{figure}

By analyzing the activations of neurons in the DRL network, we observe that after task learning, the activations of certain neurons are strongly correlated with the task goal. For instance, Figure \ref{fig:activation} illustrates the activation distribution of a specific neuron in the network following training on the drawer-open task from the Meta-World benchmark\cite{meta-world}. Activation of the neuron serves as a reliable predictor of task success. Higher activation levels correspond to an increased likelihood of completing the task successfully, indicating that this neuron encodes a critical skill essential for the task. Consequently, it plays a pivotal role in retaining task-specific memory.

Motivated by the aforementioned observations, in this work, we tackle the stability-plasticity dilemma from the perspective of neurons, and propose \textbf{Neuron-level Balance between Stability and Plasticity (NBSP)}, a novel DRL framework that operates at the level of individual neurons. 
In particular, (1) we first introduce \textbf{RL skill neurons}, which encode critical skills necessary for knowledge retention. While skill neurons have been extensively investigated and successfully exploited in various domains, such as pre-trained language models~\citep{skill_neuron} and neural machine translation~\citep{nmt_neuron}, skill neurons are still much less explored in DRL. We bridge this research gap by proposing a goal-oriented strategy for identifying RL skill neurons. (2) We then apply \textbf{gradient masking} to these neurons, ensuring that the encoded knowledge of prior skills is preserved while allowing fine-tuning during subsequent training. Meanwhile, the other neurons retain the ability needed to learn new tasks. (3) Additionally, we incorporate \textbf{experience replay} to periodically revisit the past experience to reinforce stability, preventing excessive drift from previously acquired knowledge.
Integrally, NBSP offers three key advantages compared with previous methods: (1) The neuron-level processing enables finer control and greater flexibility, addressing the stability-plasticity trade-off at the most fundamental level of the network. (2) The goal-oriented approach for identifying RL skill neurons is specifically tailored to DRL. (3) This framework is simple and parameter-free, avoiding complex network designs or additional parameters. 

We conduct experiments on the \textbf{Meta-World} \citep{meta-world} and \textbf{Atari} \citep{atari} benchmarks to evaluate the effectiveness of NBSP. Our results show that NBSP outperforms existing methods in balancing stability and plasticity, enabling effective learning of new tasks while preserving knowledge from previous tasks. Additionally, we perform extensive ablation studies to investigate the contribution of different components within NBSP. Specially, we analyze the DRL agents by dissecting the performance of the two critical modules, i.e., the actor and the critic, to assess their contributions in balancing stability and plasticity. Our findings reveal that (1) addressing both the actor and critic networks yields the best performance, and (2) the critic plays a more critical role in achieving this balance due to the differences in their inherent training mechanisms. 

In summary, our key contributions include:
\vspace{-1em}
\begin{itemize}
    \item We are the first to introduce the concept of RL skill neurons which encode skills of the task, essential for knowledge retention, and propose a goal-oriented strategy specifically tailored to DRL for identification.
    \item We tackle the stability-plasticity dilemma in DRL from the perspective of RL skill neurons, by employing gradient masking and experience replay on these neurons, eliminating requirements of complex network designs or additional parameters.
    \item We conduct extensive experiments on the Meta-World and Atari benchmarks to demonstrate the effectiveness of our method in balancing stability and plasticity. 
    % Furthermore, we highlight the crucial role of the critic in achieving this balance, shedding insights on future research.
\end{itemize}

\vspace{-1em}
\section{Related Work}
\textbf{Balance between stability and plasticity}.
In DRL, addressing the stability-plasticity dilemma \citep{Carpenter_Grossberg} has inspired various strategies. Stability-focused methods, such as replay-based approaches, including A-GEM \citep{A-GEN}, using episodic memory to constrain loss, and ClonEx-SAC \citep{clone-sac}, enhancing performance through actor cloning and exploration. Pseudo-rehearsals from generative models further reduce storage requirements \citep{Pseudo-rehearsal}. Plasticity-focused methods aim to preserve network expressiveness, with solutions like CBP \citep{dohare2024loss}, resetting \citep{primacy_bias}, plasticity injection \citep{plasticity_injection}, Reset \& Distillation \citep{replay}, and CRelu \citep{plasticity2} to prevent activation collapse. Modularity-based methods balance stability and plasticity by decoupling task-specific knowledge, exemplified by soft modularity for routing networks \citep{SM}, value function decomposition \citep{modularity1}, and compositional frameworks leveraging neural components \citep{NC}. 
Methods such as CRelu and ClonEx-SAC focus on continual reinforcement learning(CRL), but our study specifically targets the intrinsic balance between stability and plasticity with other factors such as task order controlled. We follow a cycling task setup, which allows us to assess the ability of agents to retain knowledge when revisiting previously learned tasks. Moreover, while most methods operate at the network level, our approach explores neuron-level research, providing fine-grained control.

\textbf{Neuron-level research}.
% Recent research has revealed that neuron sparsity often correlates positively with task-specific performance \citep{roles}, leading to a growing focus on identifying and leveraging skill neurons to interpret network behavior and address challenges across domains. For instance, skill neurons have been utilized to enhance transferability and efficiency in Transformers through pruning \citep{skill_neuron}, while dormant neurons have been recycled to improve training in deep reinforcement learning \citep{dormant_neuron}. Other works, such as identifying Rosetta Neurons \citep{rosetta_neuron}, language-specific neurons \citep{Language-specific_neurons} have advanced alignment and interpretability. However, in DRL, neuron-level studies remain scarce, with approaches like CoTASP \citep{CoTASP} and PackNet \citep{Packnet} focusing on task-specific sub-network selection through sparse prompts, pruning, and re-training. And NPC \citep{important_neuron} restricts important neurons to maintain stability. Distinct from these, our work identifies RL skill neurons tailored to DRL, preserving task-relevant knowledge within these neurons while allowing fine-tuning to maintain adaptability for learning other tasks.
Recent research has shown that neuron sparsity often correlates with task-specific performance \citep{roles}, driving a growing focus on skill neurons to interpret network behavior and tackle challenges across domains. For example, skill neurons have been used to enhance transferability and efficiency in Transformers via pruning \citep{skill_neuron}, and dormant neurons have been recycled to improve training in DRL\citep{dormant_neuron}. Other studies, such as identifying Rosetta Neurons \citep{rosetta_neuron} and language-specific neurons \citep{Language-specific_neurons}, have advanced alignment and interpretability. However, neuron-level studies in DRL are still limited, with methods like CoTASP \citep{CoTASP} and PackNet \citep{Packnet} focusing on task-specific sub-network selection via sparse prompts, pruning, and re-training. NPC \citep{important_neuron} restricts important neurons to maintain stability. In contrast, our work identifies RL skill neurons specific to DRL, preserving task-relevant knowledge within these neurons while allowing fine-tuning to retain adaptability for learning new tasks.

\section{Balance between Stability and Plasticity}

In this section, we first introduce the terminology of RL skill neurons and then propose the Neuron-level Balance between Stability and Plasticity (NBSP) method.

% \ls{the problem steup is not clear. Which problem do we want to solve? Here, we may add a problem setup subsection to show the settings and notations}
\subsection{Problem Setup}
In DRL, agents learn a sequence of tasks $\tau\in \{\tau_1, \tau_2, ...\}$ continually, each task $\tau$ corresponds to a distinct Markov Decision Process (MDP) $M^\tau = (S^\tau, A^\tau, P^\tau, R^\tau, \gamma^\tau )$, where $S^\tau$, $A^\tau$, $P^\tau$, $R^\tau$ and $\gamma^\tau$ denote the state space, action space, transition dynamics, reward function, and discount factor, respectively. Instead of addressing a single MDP, the goal is to solve a sequence of MDPs one by one using a universal policy $\pi(a|s)$ and Q-function $Q(s,a)$. The primary challenge lies in achieving a balance between plasticity and stability. Specifically, plasticity refers to maximizing the discounted return of the current task, while stability emphasizes the maximization of the
expected discounted return averaged across all previous tasks. How to balance this trade-off is the main problem we study in this work.

\subsection{Identifying RL Skill Neurons}

\begin{figure*}[t]
% \centering
\begin{center}
%\framebox[4.0in]{$\;$}
\vspace{0.5 em}
\includegraphics[width=\linewidth]{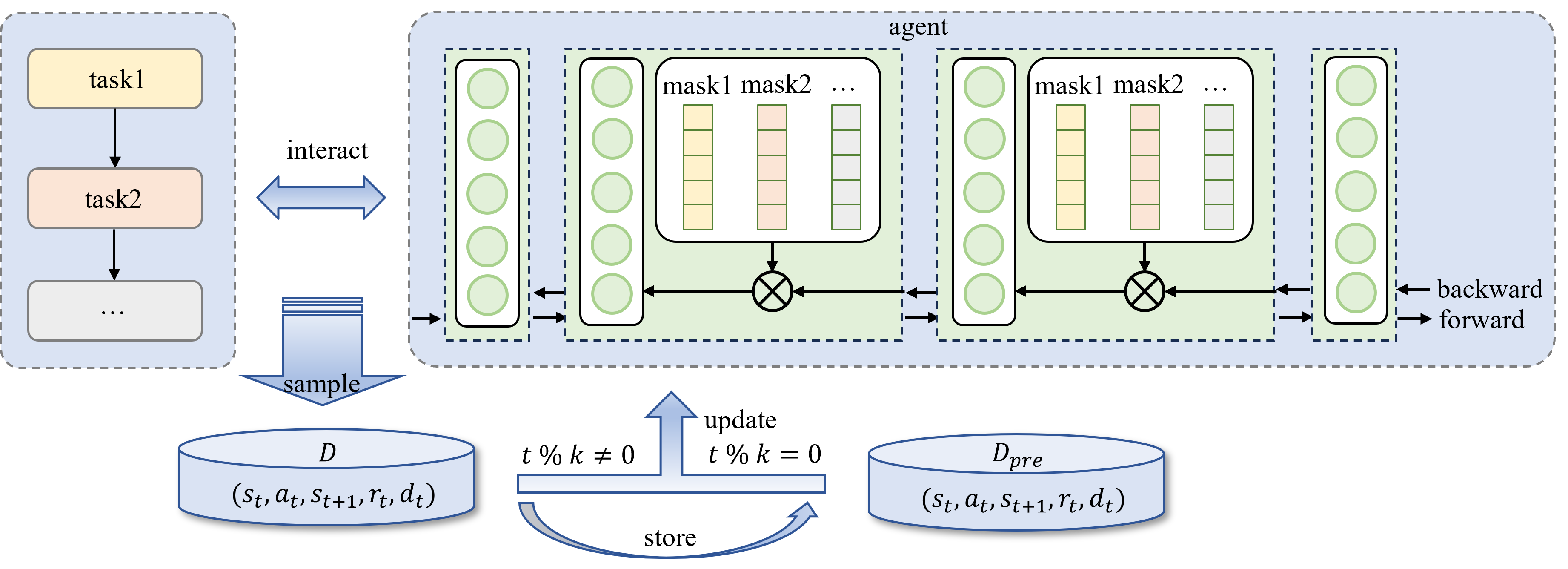}
\end{center}
\vspace{-0.5 em}
\caption{Framework of NBSP. The agent scores and identifies RL skill neurons for each task. While learning new tasks, the gradient of these neurons is masked based on their scores to preserve the encoded skills, while still allowing fine-tuning for new task learning. Additionally, a replay buffer is used to store a portion of the experiences from previous tasks, which is periodically sampled to update the agent, ensuring that knowledge from earlier tasks is retained.}
\label{fig:nbsp}
\vspace{-0.5 em}
\end{figure*}

In this study, we make a key observation that the stability and plasticity of the agent network are closely related to its expressive capabilities, which are significantly influenced by the behavior of individual neurons. As evidenced in \citet{expression}, neuron expression determines how information is propagated and processed within the neural network, directly affecting the learning and knowledge retention capabilities of the network. Therefore, understanding and controlling neuron behavior is at the most fundamental level for striking a balance between stability and plasticity.
On the one hand, when neuron expression is stable and generalized, the agent network tends to exhibit high stability. On the other hand, strong plasticity can be achieved given neuron expression is flexible and adaptable. 

Several works have demonstrated the multifaceted capabilities of neurons, such as the storage of factual knowledge \citep{factual_info}, the association with specific languages \citep{Language-specific_neurons}, and the encoding of safety information \citep{saftty_neuron}. These specialized neurons, often referred as skill neurons, have been shown to significantly contribute to network performance \citep{skill_neuron}. However, the potential of skill neurons in DRL remains largely under-explored. 
As illustrated in Figure \ref{fig:activation}, activations of the specific neuron are strongly correlated with task success: higher activation levels increase the likelihood of successful task completion, whereas lower levels are associated with failure. \textbf{\textit{This indicates that the activations of these neurons significantly affect agent performance, effectively encoding the critical skills required for the task. By preserving the activations of such neurons, it becomes possible to retain the learned task-specific skills, thereby improving stability.}}

In this work, we formally define these special neurons as \textbf{RL skill neurons}, which encode critical skills, essential for knowledge retention in DRL. Furthermore, we propose a goal-oriented method for the identification of these neurons. Unlike prior approaches that primarily focus on the inputs triggering neuron activations \citep{input_activate1, input_activate2}, our method emphasizes their impact on achieving ultimate goals, i.e. succeeding in finishing Meta-World tasks and attaining high scores in Atari games, by comparing the activation patterns between the neurons of agents that exhibit varying performance levels. In Section~\ref{ablation_selection}, we empirically show that the proposed goal-oriented method can better identify neurons that are truly encoding task-specific RL skills.

For a specific neuron $\mathcal{N}$, let $a(\mathcal{N}, t)$ represent its activation at step $t$. In fully connected layers, each output dimension corresponds to the activation of a specific neuron, whereas in convolution layers, the average of each output channel represents the activation of a neuron.
To quantify activation level of a neuron $\mathcal{N}$, we define the \textbf{standard activation} as:
\begin{equation}
\label{eq:standard_activation}
    \overline{a}(\mathcal{N}) = \frac{1}{T}\sum_{t=1}^T a(\mathcal{N}, t),
\end{equation}
where $T$ represents the evaluation step. The activation level of the neuron can then be assessed by comparing its current activation with the standard activation.

To assess the performance of the agent at step $t$, we introduce the \textbf{Goal Proximity Metric (GPM)}, denoted as $q(t)$. This metric quantifies the extent to which the agent progresses toward the goal of task, varying based on the specific objective of the task. For instance, on the Meta-World benchmark, the GPM is typically binary, representing whether the agent successfully completes the task. In contrast, the GPM is calculated based on the return achieved over an episode for the Atari benchmark.
Additionally, we define the \textbf{standard Goal Proximity Metric} (GPM) of the agent as follows, which serves as a baseline for evaluating the performance by comparing it with the current GPM.
\begin{equation}
\label{eq:standard_GPM}
    \overline{q} = \frac{1}{T}\sum_{t=1}^T q(t).
\end{equation}

To differentiate the roles of neurons across various tasks, it is essential to assess neuron activations in relation to specific goals. Intuitively, we can consider a neuron $\mathcal{N}$ to be positively contributing to the goal at step $t$ when its activation $a(\mathcal{N},t)$ surpasses the standard activation $\overline{a}(\mathcal{N})$, i.e. $a(\mathcal{N},t) > \overline{a}(\mathcal{N})$, while the GPM at the same step also exceeds its standard, i.e. $q(t) > \overline{q}$. To quantify this contribution, we accumulate a batch of results and define the positive accuracy as follows:

\begin{equation}
\label{eq:acc}
Acc(\mathcal{N})=\frac{\sum_{t=1}^T 1_{[1_{[a(\mathcal{N},t)>\overline{a}(\mathcal{N})]} = 1_{[q(t) > \overline{q}]}]}}{T}.
\end{equation}

Here, $1_{[condition]} \in \{0, 1\}$ denotes the indicator function, which returns $1$ if and only if the specified condition is satisfied. While Eq.~(\ref{eq:acc}) assesses the positive contribution of neurons towards achieving the goal, where higher accuracy implies a greater significance of the neuron in producing better outcome, however, it overlooks neurons that exhibit a negative correlation with the goal but still carry valuable task-related knowledge. Specifically, when the activation of a neuron falls below its standard activation, the agent performs well conversely. To this end, we define a \textbf{comprehensive score} $\mathbf{Score(\mathcal{N})}$ for the neuron that takes into account both positive and negative effects:
\begin{equation}
\label{eq:score}
    Score(\mathcal{N}) = max(Acc(\mathcal{N}), 1-Acc(\mathcal{N})).
\end{equation}
Subsequently, we rank all neurons in the agent network, excluding those in the last layer, in descending order based on their scores. The neurons with the highest scores are identified as RL skill neurons, as they are instrumental in retention of task-specific knowledge. The algorithm of the identification method is shown in Appendix \ref{algorithm}.

\subsection{Neuron-level Balance between Stability and Plasticity}

Building upon the concept of RL skill neurons, we propose a novel DRL framework --- \textbf{Neuron-level Balance between Stability and Plasticity (NBSP)}, as shown in Figure \ref{fig:nbsp}. Unlike prior methods \citep{modularity2, aux}, the framework proposed does not require complex network designs or additional parameters. 
Given that RL skill neurons encode essential task-specific skills, preserving their activation patterns is critical to maintaining knowledge from previous tasks during continual tasks learning. However, simply freezing RL skill neurons would hinder the ability of the agent to adapt to new tasks. To address this challenge, NBSP employs a \textbf{gradient masking} technique to balance stability and plasticity. Specifically, during each training update in the continual learning process, the gradients of RL skill neurons are selectively masked to restrict changes in their activation patterns while allowing other neurons to adapt freely. This process is formally expressed as follows:
\begin{equation}
\Delta W^{\prime(l)}_{:,j} = mask^{(l)}_j \cdot \Delta W^{(l)}_{:,j},
\end{equation}
where $\Delta W^{(l)}_{:,j}$ denotes the gradient with respect to the weight $W^{(l)}_{:,j}$ in the $l$-th layer of the network, and $j$ is the index of the output neuron in that layer. The term $mask^{(l)}_j$ is associated with the score of $j$-th neuron in the $l$-th layer, which could be calculate as follows:
\begin{equation}
\label{eq:mask}
\small
mask(\mathcal{N}) = \begin{cases}   \alpha(1- Score(\mathcal{N})) &\text{ if } \mathcal{N} \in  \{\mathcal{N}_{RL\,skill}\} \\   1 &\text{ if } \mathcal{N} \notin  \{\mathcal{N}_{RL\,skill}\}\end{cases},
\end{equation}
where $\{\mathcal{N}_{RL\,skill}\}$ represents the set of RL skill neurons, and $\alpha$ is a parameter that determines the degree of restriction on the activation pattern of these neurons, which is configured to 0.2 in the experiment. \textbf{\textit{By employing gradient masking, NBSP effectively safeguards the encoded skills within RL skill neurons from interference during the learning of new tasks, thereby enhancing stability. At the same time, RL skill neurons remain adaptable, allowing fine-tuning to accommodate new tasks and maintaining high plasticity. In addition, neurons except RL skill neurons are free to fully engage in learning new task-specific knowledge, ensuring comprehensive learning across tasks.}}

To mitigate excessive drift from knowledge acquired in previous tasks, we integrate the \textbf{experience replay} technique, periodically sampling prior experiences at specific intervals $k$. After training on a task, a portion of the experiences, rather than the entirety, are stored in a unified replay buffer $D_{pre}$, requiring only a modest memory footprint. By incorporating experience replay, the stability of DRL agents is further enhanced. The corresponding loss function is defined as follows:
\begin{equation}
\begin{aligned}
\mathcal{L}  &= R(t) \cdot \mathbb{E}_{(s_t, a_t, s_{t+1}, r_t, d_t) \sim D_{pre}}[L] \\ &+ (1-R(t)) \cdot \mathbb{E}_{(s_t, a_t, s_{t+1}, r_t, d_t) \sim D}[L],
\end{aligned}
\end{equation}
where $L$ denotes the original loss function, $R(t)$ is a binary function that evaluates to $1$ if and only if the current step $t$ is at an interval. $D$ represents the replay buffer for the current task, and $(s_t, a_t, s_{t+1}, r_t, d_t)$ denotes the tuple of the current state, action, next state, reward, and whether the episode is done sampled from the replay buffer.
The overall algorithm of NBSP is presented in Appendix \ref{algorithm}.

% In this study, we implement NBSP within the Soft Actor-Critic (SAC) framework \citep{sac}. The overall architecture is illustrated in Figure \ref{fig:nbsp}. During the learning of the second task, the gradients of RL skill neurons within the actor and critic networks are masked to retain the knowledge acquired from the previous task. Additionally, we use two separate replay buffers for experience replay: one for storing the current experiences and the other for preserving experiences from the previous task. The agent then selectively samples from these buffers to update networks.
\section{Experiment}
In this section, we evaluate the performance of NBSP on the \textbf{Meta-World} \citep{meta-world} and \textbf{Atari} benchmarks \citep{atari}.

\begin{table*}[htbp]
\centering
\caption{Results of NBSP with other baselines on the Meta-World benchmark.}
\label{tab:results_meta-world}
\vspace{0.5em}
\renewcommand{\arraystretch}{1} % 调整行高
\resizebox{\textwidth}{!}{%
\begin{tabular}{>{\centering\arraybackslash}m{4cm} >{\centering\arraybackslash}m{1.5cm} cccccccc}
\toprule
\multirow{2}{*}{\textbf{Cycling sequential tasks}} & 
\multirow{2}{*}{\textbf{Metrics}} &
\multicolumn{8}{c}{\textbf{Methods}} \\ 
\cmidrule(lr){3-10}
 &  & \textbf{EWC} & \textbf{NPC} & \textbf{ANCL} & \textbf{CoTASP} & \textbf{CRelu} & \textbf{CBP} & \textbf{PI} & \textbf{NBSP} \\ \midrule
\multirow{3}{4cm}{\centering \textbf{(window-open \\ $\rightarrow$ window-close)}} 
 & \textbf{ASR $\uparrow$} & 0.63 $\pm$ 0.03 & 0.26 $\pm$ 0.01 & 0.66 $\pm$ 0.04 & 0.05 $\pm$ 0.01  & 0.26 $\pm$ 0.14 & 0.67 $\pm$ 0.05  & 0.61 $\pm$ 0.02 & \textbf{ 0.90 $\pm$ 0.04 } \\ \cmidrule(lr){2-2}
 & \textbf{FM $\downarrow$} & 0.89 $\pm$ 0.07 & 0.68 $\pm$ 0.04 & 0.84 $\pm$ 0.10 & 0.01 $\pm$ 0.01 & 0.66 $\pm$ 0.42 & 0.78 $\pm$ 0.13  & 0.91 $\pm$ 0.07 & \textbf{ 0.18 $\pm$ 0.01 } \\ \cmidrule(lr){2-2}
 & \textbf{FWT $\uparrow$} & 0.97 $\pm$ 0.02 & 0.26 $\pm$ 0.01 & 0.97 $\pm$ 0.03 & 0.04 $\pm$ 0.01 & 0.33 $\pm$ 0.19 & 0.95 $\pm$ 0.02  & 0.95 $\pm$ 0.01 & \textbf{ 0.96 $\pm$ 0.02 } \\ \midrule
\multirow{3}{4cm}{\centering \textbf{(drawer-open \\ $\rightarrow$ drawer-close)}} 
 & \textbf{ASR $\uparrow$} & 0.68 $\pm$ 0.06 & 0.35 $\pm$ 0.05 & 0.64 $\pm$ 0.02 & 0.07 $\pm$ 0.01 & 0.29 $\pm$ 0.20  & 0.61 $\pm$ 0.03  & 0.60 $\pm$ 0.07 & \textbf{ 0.96 $\pm$ 0.02 } \\ \cmidrule(lr){2-2}
 & \textbf{FM $\downarrow$} & 0.80 $\pm$ 0.15 & 0.69 $\pm$ 0.05 & 0.88 $\pm$ 0.09 & 0.01 $\pm$ 0.01 & 0.31 $\pm$ 0.32  & 0.91 $\pm$ 0.03 & 0.71 $\pm$ 0.30 & \textbf{ 0.07 $\pm$ 0.06 } \\ \cmidrule(lr){2-2}
 & \textbf{FWT $\uparrow$} & 0.98 $\pm$ 0.01 & 0.39 $\pm$ 0.09 & 0.96 $\pm$ 0.01 & 0.09 $\pm$ 0.00  & 0.42 $\pm$ 0.28  & 0.93 $\pm$ 0.04 & 0.88 $\pm$ 0.15 & \textbf{ 0.98 $\pm$ 0.01 } \\ \midrule
\multirow{3}{4cm}{\centering \textbf{(button-press-topdown \\ $\rightarrow$ window-open)}} 
 & \textbf{ASR $\uparrow$} & 0.66 $\pm$ 0.06 & 0.25 $\pm$ 0.00 & 0.61 $\pm$ 0.01 & 0.03 $\pm$ 0.00 & 0.33 $\pm$ 0.10 & 0.62 $\pm$ 0.01  & 0.63 $\pm$ 0.02 & \textbf{ 0.95 $\pm$ 0.05 } \\ \cmidrule(lr){2-2}
 & \textbf{FM $\downarrow$} & 0.85 $\pm$ 0.14 & 0.67 $\pm$ 0.00 & 0.95 $\pm$ 0.05 & 0.01 $\pm$ 0.00 & 0.94 $\pm$ 0.01 & 0.97 $\pm$ 0.03  & 0.97 $\pm$ 0.05 & \textbf{ 0.08 $\pm$ 0.12 } \\ \cmidrule(lr){2-2}
 & \textbf{FWT $\uparrow$} & 0.96 $\pm$ 0.01 & 0.25 $\pm$ 0.01 & 0.95 $\pm$ 0.03 & 0.04 $\pm$ 0.01 & 0.42 $\pm$ 0.20 & 0.98 $\pm$ 0.02  & 0.98 $\pm$ 0.02 & \textbf{ 0.98 $\pm$ 0.01 } \\ \midrule
\multirow{3}{4cm}{\centering \textbf{(window-open \\ $\rightarrow$ window-close \\ $\rightarrow$ drawer-open \\ $\rightarrow$ drawer-close)} }
 & \textbf{ASR $\uparrow$} & 0.44 $\pm$ 0.05 & 0.19 $\pm$ 0.04 & 0.48 $\pm$ 0.04 & 0.04 $\pm$ 0.01 & 0.10 $\pm$ 0.06  & 0.43 $\pm$ 0.03 & 0.41 $\pm$ 0.06 & \textbf{ 0.66 $\pm$ 0.14 } \\ \cmidrule(lr){2-2}
 & \textbf{FM $\downarrow$} & 0.74 $\pm$ 0.11 & 0.50 $\pm$ 0.02 & 0.80 $\pm$ 0.04 & 0.04 $\pm$ 0.01 & 0.39 $\pm$ 0.02  & 0.91 $\pm$ 0.05 & 0.84 $\pm$ 0.05 & \textbf{ 0.48 $\pm$ 0.18 } \\ \cmidrule(lr){2-2}
 & \textbf{FWT $\uparrow$} & 0.83 $\pm$ 0.10 & 0.20 $\pm$ 0.05 & 0.89 $\pm$ 0.06 & 0.08 $\pm$ 0.01 & 0.13 $\pm$ 0.10 & 0.97 $\pm$ 0.02 & 0.82 $\pm$ 0.10 & \textbf{ 0.89 $\pm$ 0.12 } \\ \midrule
\multirow{3}{4cm}{\centering \textbf{(button-press-topdown \\ $\rightarrow$ window-close \\ $\rightarrow$ door-open \\ $\rightarrow$ drawer-close)}} 
 & \textbf{ASR $\uparrow$} & 0.43 $\pm$ 0.03 & 0.17 $\pm$ 0.01 & 0.44 $\pm$ 0.03 & 0.04 $\pm$ 0.01 & 0.14 $\pm$ 0.11 & 0.41 $\pm$ 0.02 & 0.38 $\pm$ 0.01 & \textbf{ 0.74 $\pm$ 0.07 } \\ \cmidrule(lr){2-2}
 & \textbf{FM $\downarrow$} & 0.81 $\pm$ 0.09 & 0.47 $\pm$ 0.01 & 0.87 $\pm$ 0.02  & 0.04 $\pm$ 0.00 & 0.62 $\pm$ 0.16 & 0.94 $\pm$ 0.02 & 0.97 $\pm$ 0.02 & \textbf{ 0.34 $\pm$ 0.15 } \\ \cmidrule(lr){2-2}
 & \textbf{FWT $\uparrow$} & 0.88 $\pm$ 0.10 & 0.19 $\pm$ 0.02 & 0.91 $\pm$ 0.08 & 0.07 $\pm$ 0.02 & 0.17 $\pm$ 0.15 & 0.97 $\pm$ 0.01 & 0.92 $\pm$ 0.07 & \textbf{ 0.95 $\pm$ 0.06 } \\ 
\bottomrule
\end{tabular}%
}
\end{table*}

\begin{figure*}[t]
% \centering
\begin{center}
%\framebox[4.0in]{$\;$}
% \vspace{0.5em}
\includegraphics[width=\linewidth]{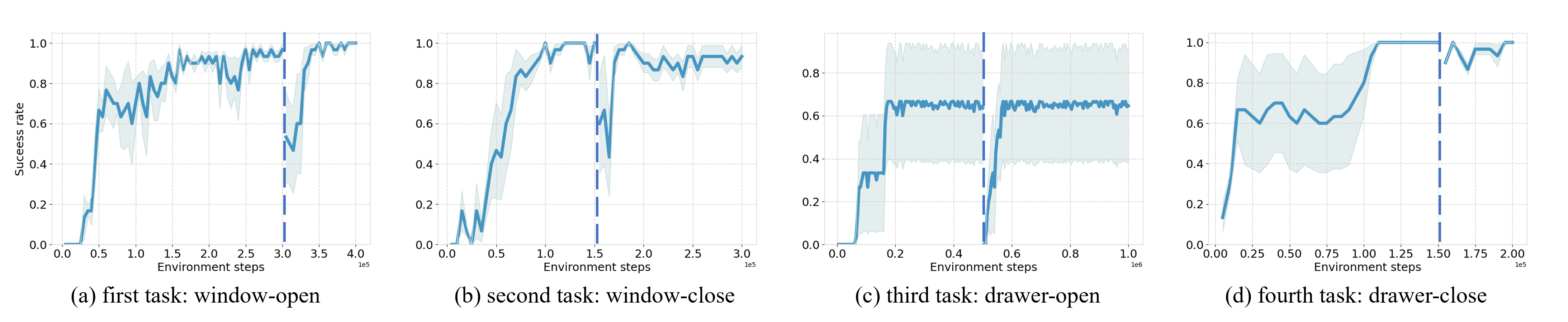}
\end{center}
\vspace{-1.5em}
\caption{Training process of NBSP on the Meta-World benchmark. The segments to the left and right of the dashed line represent the training processes of the first and second cycles, respectively.}
\label{fig:training}
\vspace{-0.5em}
\end{figure*}

\textbf{Experiment setting}.
We follow the the experimental paradigm of \citet{plasticity2, neuroplastic}, evaluating our proposed method on a \textbf{cycling sequence of tasks}
characterized by non-stationarity due to changing environments over time. Specifically, the agent learns each task sequentially and transitions to the next without resetting the learned networks. The task cycles through a fixed sequence, with a cycle completing once all tasks in the sequence have been learned. The agent cycles twice, resulting in each task being repeated twice during the training process.
Compared to the CRL training paradigm, our cycling training paradigm provides a more specific evaluation of the balance between stability and plasticity. By repeating each task twice within a cycling sequence, the setup not only assesses the plasticity in adapting to new tasks but also evaluates its stability when revisiting previously learned tasks, avoiding the influence of task order.
For Meta-World benchmark, experiments are conducted on three groups of two-task cycling tasks and two groups of four-task cycling tasks. For Atari, experiments are conducted on two groups of two-game cycling tasks. Details about the benchmarks are shown in Appendix \ref{benchmark}.

For all experiments, we use the Soft Actor-Critic (SAC)~\citep{sac} algorithm, as implemented by CleanRL \citep{cleanrl}. Each agent is trained until either reaching a predefined maximum number of steps or demonstrating stable mastery of the task in the Meta-World benchmark. To ensure the robustness of our results, each experiment is repeated using three different random seeds. Detailed descriptions of the hyperparameters and other experimental settings are provided in Appendix \ref{setting}.

\textbf{Metric}.
Overall performance is commonly assessed using the \textbf{Average Success Rate (ASR)}, analogous to the Average Incremental Accuracy (AIA) metric \citep{AIA}. Let $sr_{i,j}$ represents the success rate evaluated on the $j$-th task after completing the learning of the $i$-th task ($i \ge j$). The ASR is then defined as:
\begin{equation}
    ASR = \frac{1}{k}\sum_{i=1}^{k}{\frac{1}{i}\sum_{i \ge j}{sr_{i,j}}},
\end{equation}
where $k$ represents the number of tasks. The higher the ASR, the better the method balances stability and plasticity.

To evaluate the stability of the agent, we utilize the \textbf{Forgetting Measure (FM)}~\citep{FM}. The lower the FM, the better the method maintains stability. In our experiments, the FM is calculated as follows:
\begin{equation}
\small
    FM = \frac{1}{k-1}\sum_{i=2}^{k}{\frac{1}{i-1}\sum_{i\ge j}\mathop{max} \limits_{l\in\{1,...,i-1\}} (sr_{l,j}-sr_{i,j})}.
\end{equation}

To assess the plasticity of the agent, we employ the \textbf{Forward Transfer (FWT)} metric~\citep{FWT}, which is calculated as follows:
\begin{equation}
    FWT = \frac{1}{k}\sum_{i=1}^k sr_{i,i}.
\end{equation}
The higher the FWT, the better the method maintains plasticity. Further details about evaluation metrics are available in Appendix \ref{metric}.

\textbf{Baseline}. 
To assess the effectiveness of our proposed NBSP framework, we compare it with seven baseline methods dealing with the balance between stability and plasticity. \textbf{EWC} \citep{ewc} and \textbf{NPC} \citep{important_neuron} primarily emphasize maintaining stability, while \textbf{CRelu} \citep{plasticity2}, \textbf{CBP} \citep{dohare2024loss}, and \textbf{PI} \citep{plasticity_injection} focus on enhancing plasticity. \textbf{ANCL} \citep{aux} and \textbf{CoTASP} \citep{CoTASP} aim to achieve a balance between stability and plasticity. Notably, CoTASP makes relevant tasks share more neurons in the meta-policy network, and NPC estimates the importance value of each neuron and consolidates important neurons, they are both relevant to neurons. Detailed descriptions of these baselines can be found in Appendix \ref{baseline}.

\subsection{Experiment on the Meta-World Benchmark}
\label{experiment_meta-world}

The experimental results of NBSP compared with other baselines on the Meta-World benchmark are presented in Table \ref{tab:results_meta-world}. As shown in the final column, NBSP significantly outperforms all other methods across evaluation metrics, including ASR, FM, and FWT. For two-task cycling tasks, NBSP achieves an ASR consistently above 0.9, which is substantially higher than other baselines. Its stability metric, FM, is markedly lower, while its plasticity metric, FWT, remains at a high level. Furthermore, NBSP also demonstrates excellent performance in four-task cycling tasks, maintaining a substantial lead over all baselines.

For stability-focused baselines, EWC achieves a relatively good ASR compared to other baselines but still falls short of NBSP. Moreover, EWC exhibits poor stability due to its high FM values. NPC performs even worse, failing to maintain both stability and plasticity effectively. Among plasticity-focused baselines, CBP and PI achieve comparable plasticity to NBSP, as reflected in their high FWT scores. However, both suffer from severe stability loss, indicated by their higher FM values. Another plasticity-focused method, CRelu, underperforms in both stability and plasticity.
For baselines attempting to balance stability and plasticity, ANCL maintains high plasticity with competitive FWT scores but struggles to retain prior knowledge, as indicated by its poor FM performance. CoTASP, designed for balancing stability and plasticity, performs poorly overall.

The effectiveness of NBSP is further demonstrated in Figure \ref{fig:training}, which showcases the training dynamics of NBSP. Specifically, during the second cycle of learning the same task, the agent exhibits a high success rate even before retraining, indicating that it has retained significant task knowledge. As a result, the agent is able to master the task more rapidly. This highlights the ability of NBSP to preserve knowledge from prior tasks while simultaneously maintaining the plasticity required to learn new tasks effectively. The other training process is demonstrated in Appendix \ref{result_meta-world}. In summary,
\textbf{\textit{NBSP delivers a remarkable improvement in maintaining stability without compromising plasticity, achieving a well-balanced trade-off in DRL.}}

\subsection{Ablation Study}
\label{ablation}

\begin{table}[t]
\centering
\renewcommand{\arraystretch}{1} % 调整行高
\caption{Results of ablation study of gradient masking and experience replay techniques.}
\label{tab:ablation_component}
\vspace{0.5em}
\resizebox{\columnwidth}{!}{%
\begin{tabular}{>{\centering\arraybackslash}m{1.8cm} >{\centering\arraybackslash}m{1.8cm}ccc}
\toprule
\multicolumn{2}{c}{\textbf{Components}} & \multicolumn{3}{c}{\textbf{Metrics}} \\ 
\cmidrule(lr){1-2} \cmidrule(lr){3-5}
\textbf{Gradient Masking} & \textbf{Experience Replay} & \textbf{ASR $\uparrow$} & \textbf{FM $\downarrow$} & \textbf{FT $\uparrow$} \\ 
\midrule
$\times$   & $\times$   & 0.62 $\pm$ 0.01   & 0.99 $\pm$ 0.02   & 0.98 $\pm$ 0.02 \\ 
$\times$   & \checkmark & 0.70 $\pm$ 0.08   & 0.50 $\pm$ 0.16   & 0.92 $\pm$ 0.05 \\ 
\checkmark & $\times$   & 0.71 $\pm$ 0.06   & 0.73 $\pm$ 0.21   & 0.97 $\pm$ 0.02\\ 
\checkmark & \checkmark & \textbf{ 0.95 $\pm$ 0.05 }  & \textbf{ 0.08 $\pm$ 0.12 }  & \textbf{ 0.98 $\pm$ 0.01 } \\ 
\bottomrule
\end{tabular}%
}
\vspace{-1em}
\end{table}

\begin{table}[t]
\centering
\renewcommand{\arraystretch}{1} % 调整行高
\caption{Results of ablation study of neuron identification methods.}
\label{tab:ablation_selection}
\vspace{0.5em}
\resizebox{\columnwidth}{!}{%
\begin{tabular}{ >{\centering\arraybackslash}m{4cm} >{\centering\arraybackslash}m{1.8cm} ccc}
\toprule

\multirow{2}{*}{\textbf{Cycling sequential tasks}} &
\multirow{2}{*}{\textbf{Methods}} &
\multicolumn{3}{c}{\textbf{Metrics}} \\ 
\cmidrule(lr){3-5}
& & \textbf{ASR $\uparrow$} & \textbf{FM $\downarrow$} & \textbf{FT $\uparrow$} \\ 
\midrule

\multirow{2}{4cm}{\centering \textbf{(window-open \\ $\rightarrow$ window-close)}} 
& \textbf{random}     & 0.78 $\pm$ 0.09   & 0.42 $\pm$ 0.13   & 0.90 $\pm$ 0.06 \\ 
\cmidrule(lr){2-2}
& \textbf{ours}        & \textbf{0.90 $\pm$ 0.04}   & \textbf{0.18 $\pm$ 0.01}   & \textbf{0.96 $\pm$ 0.02} \\ 
\midrule

\multirow{2}{4cm}{\centering \textbf{(drawer-open \\ $\rightarrow$ drawer-close)}} 
& \textbf{random}     & 0.72 $\pm$ 0.26   & 0.41 $\pm$ 0.28   & 0.83 $\pm$ 0.23 \\ 
\cmidrule(lr){2-2}
& \textbf{ours}        & \textbf{0.96 $\pm$ 0.02}   & \textbf{0.07 $\pm$ 0.06}   & \textbf{0.98 $\pm$ 0.01} \\ 
\midrule

\multirow{2}{4cm}{\centering \textbf{(button-press-topdown \\ $\rightarrow$ window-open)}} 
& \textbf{random}     & 0.72 $\pm$ 0.01   & 0.70 $\pm$ 0.05   & 0.96 $\pm$ 0.02 \\ 
\cmidrule(lr){2-2}
& \textbf{ours}        & \textbf{0.95 $\pm$ 0.05}   & \textbf{0.08 $\pm$ 0.12}   & \textbf{0.98 $\pm$ 0.01} \\ 
\bottomrule

\end{tabular}%
}
\vspace{-1em}
\end{table}

\begin{table}[t]
\centering
\renewcommand{\arraystretch}{1} % 调整行高
\caption{Results of ablation study of the actor and critic modules.}
\label{tab:ablation_module}
\vspace{0.5em}
\resizebox{\columnwidth}{!}{%
\begin{tabular}{ >{\centering\arraybackslash}m{4cm} >{\centering\arraybackslash}m{1.8cm} ccc}
\toprule

\multirow{2}{*}{\textbf{Cycling sequential tasks}} &
\multirow{2}{*}{\textbf{Modules}} &
\multicolumn{3}{c}{\textbf{Metrics}} \\ 
\cmidrule(lr){3-5}
& & \textbf{ASR $\uparrow$} & \textbf{FM $\downarrow$} & \textbf{FT $\uparrow$} \\ 
\midrule

\multirow{3}{4cm}{\centering \textbf{(window-open \\ $\rightarrow$ window-close)}} 
& \textbf{actor}     & 0.76 $\pm$ 0.10   & 0.58 $\pm$ 0.19   & 0.97 $\pm$ 0.04 \\ 
\cmidrule(lr){2-2}
& \textbf{critic}        & 0.79 $\pm$ 0.05   & 0.48 $\pm$ 0.09   & 0.94 $\pm$ 0.05 \\ 
\cmidrule(lr){2-2}
& \textbf{both}        & \textbf{0.90 $\pm$ 0.04}   & \textbf{0.18 $\pm$ 0.01}   & \textbf{0.96 $\pm$ 0.02} \\ 
\midrule

\multirow{3}{4cm}{\centering \textbf{(drawer-open \\ $\rightarrow$ drawer-close)}} 
& \textbf{actor}     & 0.79 $\pm$ 0.05   & 0.55 $\pm$ 0.15   & 0.99 $\pm$ 0.01 \\ 
\cmidrule(lr){2-2}
& \textbf{critic}        & 0.86 $\pm$ 0.02   & 0.31 $\pm$ 0.03   & 0.96 $\pm$ 0.02 \\ 
\cmidrule(lr){2-2}
& \textbf{both}        & \textbf{0.96 $\pm$ 0.02}   & \textbf{0.07 $\pm$ 0.06}   & \textbf{0.98 $\pm$ 0.01} \\ 
\midrule

\multirow{3}{4cm}{\centering \textbf{(button-press-topdown \\ $\rightarrow$ window-open)}} 
& \textbf{actor}     & 0.81 $\pm$ 0.11   & 0.45 $\pm$ 0.28   & 0.95 $\pm$ 0.01 \\ 
\cmidrule(lr){2-2}
& \textbf{critic}        & 0.85 $\pm$ 0.16   & 0.35 $\pm$ 0.38   & 0.95 $\pm$ 0.03 \\ 
\cmidrule(lr){2-2}
& \textbf{both}        & \textbf{0.95 $\pm$ 0.05}   & \textbf{0.08 $\pm$ 0.12}   & \textbf{0.98 $\pm$ 0.01} \\
\bottomrule
\end{tabular}%
}
\vspace{-1em}
\end{table}

In the ablation study, we further evaluate the effectiveness of (1) the two primary components of NBSP: the gradient masking technique and experience replay technique, (2) the neuron identification method, and (3) the two critical modules
of DRL: the actor and the critic. What's more, we analyze how the proportion of RL skill neurons influences the performance of NBSP.

\textbf{Gradient masking and experience replay}.
To assess the impact of the two primary components of NBSP, we designed four experimental settings: \textbf{(1) Base}: training directly without any additional techniques. \textbf{(2) Mask-Only}: training with only the gradient masking technique. \textbf{(3) Replay-Only}: training with only the experience replay technique. \textbf{(4) NBSP}: training with both two techniques.

The results of (button-press-topdown $\rightarrow$ window-open) cycling sequential tasks are shown in Table \ref{tab:ablation_component}. From the results, we observe the following: (1) The Base setting exhibits significant stability loss, as indicated by its high FM value. This result highlights the challenge of maintaining stability without specific mechanisms to preserve task knowledge. (2) Both the Mask-Only and Replay-Only settings alleviate the stability loss to some extent. This confirms the individual contributions of the gradient masking and experience replay techniques in mitigating forgetting and maintaining stability. (3) The combination of both techniques in NBSP yields superior performance, which is greatly improved compared to the use of only one. This is evidenced by significantly lower FM values (indicating enhanced stability) and high FWT values (demonstrating maintained plasticity). 
\textbf{\textit{These findings demonstrate that while each technique independently contributes to improving stability and maintaining plasticity, their synergy in NBSP is crucial for achieving optimal performance in cycling sequential learning scenarios.}} Additional results for different task settings are provided in Appendix \ref{ablation_component}.

\begin{figure}[t]
% \centering
\begin{center}
%\framebox[4.0in]{$\;$}
\includegraphics[width=\columnwidth]{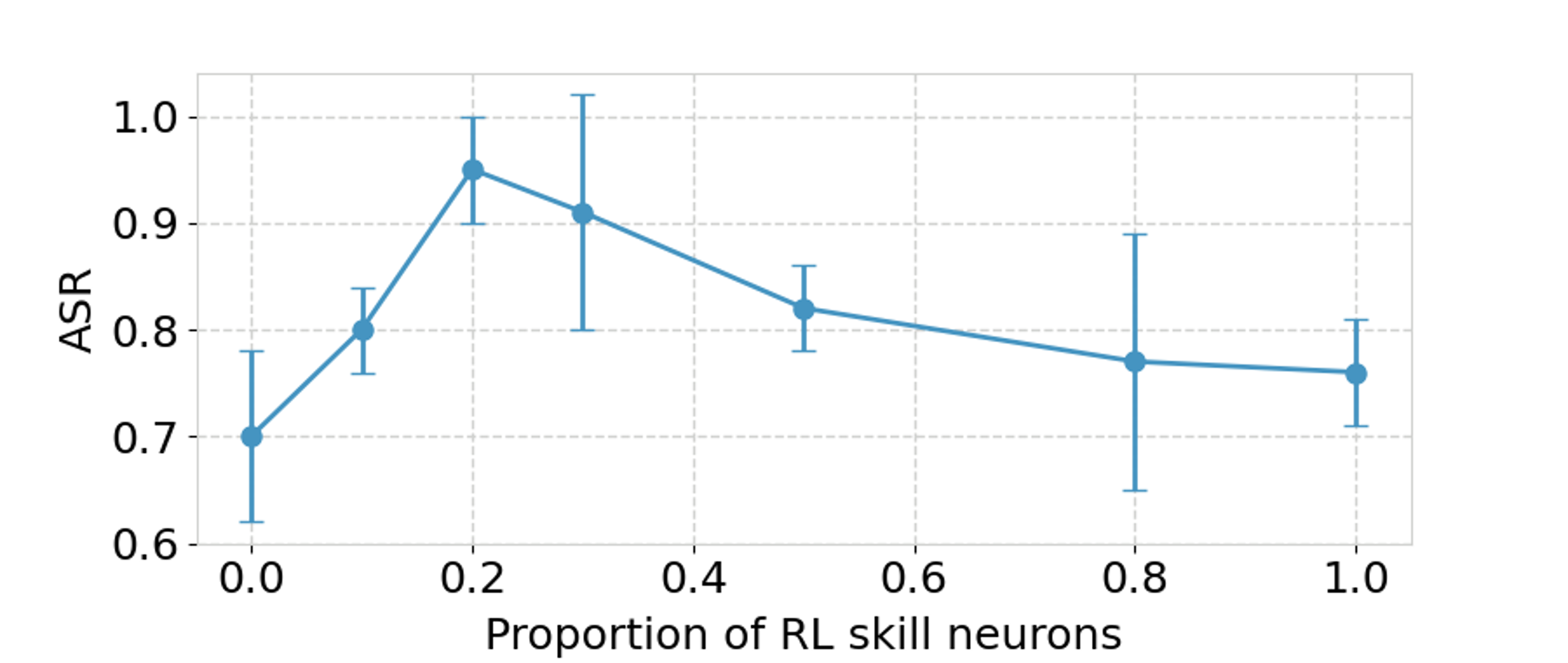}
\vspace{-2.5em}
\end{center}
\caption{Performance of NBSP with different proportions of RL skill neurons.}
\label{fig:neuron_num}
\vspace{-1em}
\end{figure}

\begin{table*}[t]
\centering
\caption{Results of NBSP with other baselines on the Atari benchmark.}
\label{tab:results_atari}
\vspace{0.5em}
\renewcommand{\arraystretch}{1} % 调整行高
\resizebox{\textwidth}{!}{%
\begin{tabular}{>{\centering\arraybackslash}m{4cm} >{\centering\arraybackslash}m{1.5cm} cccccccc}
\toprule
\multirow{2}{*}{\textbf{Cycling sequential games}} & 
\multirow{2}{*}{\textbf{Metrics}} &
\multicolumn{8}{c}{\textbf{Methods}} \\ 
\cmidrule(lr){3-10}
 &  & \textbf{EWC} & \textbf{NPC} & \textbf{ANCL} & \textbf{CoTASP} & \textbf{CRelu} & \textbf{CBP} & \textbf{PI} & \textbf{NBSP} \\ \midrule
 
\multirow{3}{4cm}{\centering \textbf{(Pong $\rightarrow$ Bowling)}} 
 & \textbf{AR $\uparrow$} & 0.66 $\pm$ 0.07 & 0.51 $\pm$ 0.02 & 0.42 $\pm$ 0.29 & -0.05 $\pm$ 0.02  &  0.02 $\pm$ 0.00  & -0.09 $\pm$ 0.00  & 0.53 $\pm$ 0.01 & \textbf{ 0.87 $\pm$ 0.01 } \\ \cmidrule(lr){2-2}
 & \textbf{FM $\downarrow$} & 0.58 $\pm$ 0.20 & 0.51 $\pm$ 0.04 & 0.46 $\pm$ 0.31 & 0.07 $\pm$ 0.01  &  0.01 $\pm$ 0.00  & 0.06 $\pm$ 0.00  & 0.78 $\pm$ 0.02 & \textbf{ 0.05 $\pm$ 0.03 } \\ \cmidrule(lr){2-2}
 & \textbf{FWT $\uparrow$} & 0.70 $\pm$ 0.02 & 0.35 $\pm$ 0.02 & 0.47 $\pm$ 0.31 & -0.05 $\pm$ 0.05  &  0.02 $\pm$ 0.01  & -0.09 $\pm$ 0.00  & 0.60 $\pm$ 0.00 & \textbf{ 0.72 $\pm$ 0.01 } \\ \midrule
 
\multirow{3}{4cm}{\centering \textbf{(BankHeist $\rightarrow$ Alien)}} 
 & \textbf{AR $\uparrow$} & 0.46 $\pm$ 0.01 & 0.38 $\pm$ 0.06 & 0.46 $\pm$ 0.01 & -0.08 $\pm$ 0.05  & 0.08 $\pm$ 0.05 & 0.12 $\pm$ 0.02  & 0.48 $\pm$ 0.14 & \textbf{ 0.57 $\pm$ 0.02 } \\ \cmidrule(lr){2-2}
 & \textbf{FM $\downarrow$} & 0.98 $\pm$ 0.02 & 0.46 $\pm$ 0.14 & 0.98 $\pm$ 0.03 &  0.27 $\pm$ 0.04  & 0.52 $\pm$ 0.29 & 0.44 $\pm$ 0.09  & 0.88 $\pm$ 0.27 & \textbf{ 0.65 $\pm$ 0.07 } \\ \cmidrule(lr){2-2}
 & \textbf{FWT $\uparrow$} & 0.71 $\pm$ 0.02 & 0.37 $\pm$ 0.03 & 0.72 $\pm$ 0.01 & -0.16 $\pm$ 0.07  & 0.28 $\pm$ 0.11 & 0.30 $\pm$ 0.05  & 0.73 $\pm$ 0.26 & \textbf{ 0.72 $\pm$ 0.05 } \\
\bottomrule
\end{tabular}%
}
\vspace{-0.5em}
\end{table*}

\textbf{Neuron identification method}. 
\label{ablation_selection}
To evaluate the proposed goal-oriented neuron identification method, we compare it with random selection. As shown in Table \ref{tab:ablation_selection}, our goal-oriented method consistently outperforms random selection across all three metrics: ASR, FM, and FWT. This result confirms that our method effectively identifies neurons critical for knowledge retention, ensuring better stability and plasticity in cycling sequential task learning. In contrast, random selection fails to prioritize essential neurons, leading to poorer overall performance. \textbf{\textit{These findings validate the necessity of task-specific, goal-oriented neuron identification in enhancing balance between stability and plasticity.}}

\textbf{Actor and critic}.
To get a deeper understanding of the individual roles of the actor and critic in DRL agents, we compare the performance of NBSP with that only applied on actor and critic.
% evaluate two variations of our method with NBSP: 
% \textbf{(1) NBSP-Actor}: applying NBSP exclusively to the neurons of the actor network;
% \textbf{(2) NBSP-Critic}: applying NBSP exclusively to the neurons of the critic network.
The result is shown in Table \ref{tab:ablation_module}.
% As illustrated, both NBSP-Actor and NBSP-Critic achieve substantial improvements in learning new tasks while retaining knowledge from the previous. Notably, NBSP-Critic performs better in retaining knowledge compared to NBSP-Actor. Despite these improvements, neither NBSP-Actor nor NBSP-Critic achieves the same level of performance as NBSP, which applies the techniques to both the actor and critic networks simultaneously.

\textbf{\textit{The results indicate that both the actor and critic networks are essential for striking an optimal balance between stability and plasticity. Notably, the critic proves to be the more critical module in balancing this trade-off}}, which aligns with the insight from \citet{visual} that plasticity loss in the critic serves as the principal bottleneck impeding efficient training in DRL. \textbf{\textit{
We further investigate this phenomenon by dissecting the inherent training mechanisms of actor-critic RL methods, and draw the following key observations}}: (1) Updates to the actor are guided by feedback from the critic. Consequently, even if the RL skill neurons in the actor are masked, they remain influenced by the critic, which may gradually adapt to the new task at the expense of retaining prior knowledge; (2) In contrast, applying NBSP to the critic network indirectly constrains the actor as well; and (3) The update process of the critic network is recursive, with its target network updated via an exponential moving average, enabling it to preserve knowledge from the previous task while integrating new skills. Therefore, NBSP achieves better performance on the critic than on the actor. This demonstrates the distinct roles of the actor and critic in balancing stability and plasticity, providing valuable insights for future research in this field.

\textbf{The proportion of RL skill neurons}.
To evaluate the impact of the proportion of RL skill neurons on the performance of NBSP, we experiment with various proportions on the (button-press-topdown $\rightarrow$ window-open) cycling tasks. The results, shown in Figure \ref{fig:neuron_num}, reveal an interesting trend: \textbf{\textit{as the proportion of RL skill neurons increases, the ASR improves initially, but begins to decline after reaching a certain threshold}}. Specifically, when the proportion is small, not all neurons encoding task-specific skills are identified, leading to knowledge loss stored in neurons that are not selected. On the other hand, when the proportion becomes too large, neurons that do not encode skills may be incorrectly selected as RL skill neurons, which compromises their learning capacity and causes the true RL skill neurons to adjust their activations to accommodate new tasks, ultimately reducing stability. Thus, determining the optimal proportion of RL skill neurons is crucial for achieving the best performance. Our experiments suggest that a proportion of 0.2 is ideal for balancing stability and plasticity.

\subsection{Experiment on the Atari Benchmark}

We further evaluate NBSP on the Atari benchmark to assess its generalization ability. In contrast to the continuous action space of Meta-World, Atari games feature discrete action spaces, and episode returns are used to evaluate the performance of each game. The results are presented in Table \ref{tab:results_atari}.
As with the Meta-World benchmark, NBSP demonstrates superior performance in balancing stability and plasticity, outperforming other baselines across key evaluation metrics, including AR (Average Return), FM, and FWT. In a word, \textit{\textbf{NBSP exhibits excellent generalization in balance stability and plasticity across different benchmarks.}}

\section{Conclusion}
% \textbf{Conclusion}. 
This work addresses the fundamental issue of the stability-plasticity dilemma in DRL. To tackle this problem, we introduce the concept of RL skill neurons by identifying neurons that significantly contribute to knowledge retention, building upon which we then propose the Neuron-level Balance between Stability and Plasticity framework, by employing gradient masking and experience replay techniques on RL skill neurons. Experimental results on the Meta-World and Atari benchmarks demonstrate that NBSP significantly outperforms existing methods in managing the stability-plasticity trade-off. Future research could explore the application of RL skill neurons like model distillation and extend NBSP to other learning paradigms, such as supervised learning.
% Furthermore, comparative studies of NBSP with its variations reveal that the critic network plays a more crucial role in achieving this balance than the actor network, which is consistent with their respective training mechanisms, shedding valuable insights on future research in this field.

% \textbf{Future work}.
% This work defines RL skill neurons, effectively balancing stability and plasticity in DRL. Future research could explore model distillation by pruning neurons apart from RL skill neurons for more compact models, and bias control by manipulating RL skill neuron activations for targeted behaviors. The NBSP method could also extend to other learning paradigms, such as supervised and unsupervised learning, to tackle similar challenges. We plan to investigate these directions further across diverse domains.

% Acknowledgements should only appear in the accepted version.
%\section*{Acknowledgements}

%\textbf{Do not} include acknowledgements in the initial version of
% the paper submitted for blind review.

\section*{Impact Statement}

% Authors are \textbf{required} to include a statement of the potential 
% broader impact of their work, including its ethical aspects and future 
% societal consequences. This statement should be in an unnumbered 
% section at the end of the paper (co-located with Acknowledgements -- 
% the two may appear in either order, but both must be before References), 
% and does not count toward the paper page limit. In many cases, where 
% the ethical impacts and expected societal implications are those that 
% are well established when advancing the field of Machine Learning, 
% substantial discussion is not required, and a simple statement such 
% as the following will suffice:
In deep reinforcement learning, the stability-plasticity dilemma refers to the challenge of balancing the retention of existing skills (stability) with the acquisition of new knowledge (plasticity). This dilemma significantly hampers the performance of agents in sequential task learning, posing obstacles for practical applications. In this work, we discover that certain neurons within the agent network play a pivotal role in shaping the agent's behavior. Leveraging this insight, we define RL skill neurons as those responsible for encoding critical task-related skills and propose a goal-oriented method to identify them. To address the stability-plasticity dilemma, we introduce gradient masking and experience replay techniques specifically targeting these neurons. These techniques preserve knowledge from previously learned tasks while allowing fine-tuning to adapt to new ones. Our proposed method, NBSP, achieves a superior balance between stability and plasticity in DRL. This study presents no ethical concerns and poses no negative impact on society.
% This paper presents work whose goal is to advance the field of 
% Machine Learning. There are many potential societal consequences 
% of our work, none which we feel must be specifically highlighted here.

% The above statement can be used verbatim in such cases, but we 
% encourage authors to think about whether there is content which does 
% warrant further discussion, as this statement will be apparent if the 
% paper is later flagged for ethics review.

% In the unusual situation where you want a paper to appear in the
% references without citing it in the main text, use \nocite
\nocite{langley00}

\bibliography{main}

\begin{thebibliography}{52}
\providecommand{\natexlab}[1]{#1}
\providecommand{\url}[1]{\texttt{#1}}
\expandafter\ifx\csname urlstyle\endcsname\relax
  \providecommand{\doi}[1]{doi: #1}\else
  \providecommand{\doi}{doi: \begingroup \urlstyle{rm}\Url}\fi

\bibitem[Abbas et~al.(2023)Abbas, Zhao, Modayil, White, and Machado]{plasticity2}
Abbas, Z., Zhao, R., Modayil, J., White, A., and Machado, M.~C.
\newblock Loss of plasticity in continual deep reinforcement learning.
\newblock In \emph{Conference on Lifelong Learning Agents}, pp.\  620--636. PMLR, 2023.

\bibitem[Ahn et~al.(2024)Ahn, Hyeon, Oh, Hwang, and Moon]{replay}
Ahn, H., Hyeon, J., Oh, Y., Hwang, B., and Moon, T.
\newblock Reset \& distill: A recipe for overcoming negative transfer in continual reinforcement learning.
\newblock \emph{arXiv preprint arXiv:2403.05066}, 2024.

\bibitem[Anand \& Precup(2024)Anand and Precup]{modularity1}
Anand, N. and Precup, D.
\newblock Prediction and control in continual reinforcement learning.
\newblock \emph{Advances in Neural Information Processing Systems}, 36, 2024.

\bibitem[Andrychowicz et~al.(2020)Andrychowicz, Baker, Chociej, Jozefowicz, McGrew, Pachocki, Petron, Plappert, Powell, Ray, et~al.]{robot}
Andrychowicz, O.~M., Baker, B., Chociej, M., Jozefowicz, R., McGrew, B., Pachocki, J., Petron, A., Plappert, M., Powell, G., Ray, A., et~al.
\newblock Learning dexterous in-hand manipulation.
\newblock \emph{The International Journal of Robotics Research}, 39\penalty0 (1):\penalty0 3--20, 2020.

\bibitem[Atkinson et~al.(2021{\natexlab{a}})Atkinson, McCane, Szymanski, and Robins]{Pseudo-rehearsal}
Atkinson, C., McCane, B., Szymanski, L., and Robins, A.
\newblock Pseudo-rehearsal: Achieving deep reinforcement learning without catastrophic forgetting.
\newblock \emph{Neurocomputing}, 428:\penalty0 291--307, 2021{\natexlab{a}}.

\bibitem[Atkinson et~al.(2021{\natexlab{b}})Atkinson, McCane, Szymanski, and Robins]{stability2}
Atkinson, C., McCane, B., Szymanski, L., and Robins, A.
\newblock Pseudo-rehearsal: Achieving deep reinforcement learning without catastrophic forgetting.
\newblock \emph{Neurocomputing}, pp.\  291–307, Mar 2021{\natexlab{b}}.
\newblock \doi{10.1016/j.neucom.2020.11.050}.
\newblock URL \url{http://dx.doi.org/10.1016/j.neucom.2020.11.050}.

\bibitem[Bai et~al.(2023)Bai, Zhang, Tao, Wu, Wang, and Xu]{modularity2}
Bai, F., Zhang, H., Tao, T., Wu, Z., Wang, Y., and Xu, B.
\newblock Picor: Multi-task deep reinforcement learning with policy correction.
\newblock In \emph{Proceedings of the AAAI Conference on Artificial Intelligence}, volume~37, pp.\  6728--6736, 2023.

\bibitem[Bau et~al.(2018)Bau, Belinkov, Sajjad, Durrani, Dalvi, and Glass]{nmt_neuron}
Bau, A., Belinkov, Y., Sajjad, H., Durrani, N., Dalvi, F., and Glass, J.
\newblock Identifying and controlling important neurons in neural machine translation.
\newblock In \emph{International Conference on Learning Representations}, 2018.

\bibitem[Bau et~al.(2020)Bau, Zhu, Strobelt, Lapedriza, Zhou, and Torralba]{input_activate1}
Bau, D., Zhu, J.-Y., Strobelt, H., Lapedriza, A., Zhou, B., and Torralba, A.
\newblock Understanding the role of individual units in a deep neural network.
\newblock \emph{Proceedings of the National Academy of Sciences}, 117\penalty0 (48):\penalty0 30071--30078, 2020.

\bibitem[Bellemare et~al.(2013)Bellemare, Naddaf, Veness, and Bowling]{ale}
Bellemare, M.~G., Naddaf, Y., Veness, J., and Bowling, M.
\newblock The arcade learning environment: An evaluation platform for general agents.
\newblock \emph{Journal of Artificial Intelligence Research}, 47:\penalty0 253--279, 2013.

\bibitem[Carpenter \& Grossberg(1987)Carpenter and Grossberg]{plasticity}
Carpenter, G.~A. and Grossberg, S.
\newblock A massively parallel architecture for a self-organizing neural pattern recognition machine.
\newblock \emph{Computer vision, graphics, and image processing}, 37\penalty0 (1):\penalty0 54--115, 1987.

\bibitem[Carpenter \& Grossberg(1988)Carpenter and Grossberg]{Carpenter_Grossberg}
Carpenter, G.~A. and Grossberg, S.
\newblock Art 2: Self-organization of stable category recognition codes for analog input patterns.
\newblock In \emph{SPIE Proceedings,Intelligent Robots and Computer Vision VI}, Feb 1988.
\newblock \doi{10.1117/12.942747}.
\newblock URL \url{http://dx.doi.org/10.1117/12.942747}.

\bibitem[Chaudhry et~al.(2018{\natexlab{a}})Chaudhry, Dokania, Ajanthan, and Torr]{FM}
Chaudhry, A., Dokania, P.~K., Ajanthan, T., and Torr, P.~H.
\newblock Riemannian walk for incremental learning: Understanding forgetting and intransigence.
\newblock In \emph{Proceedings of the European conference on computer vision (ECCV)}, pp.\  532--547, 2018{\natexlab{a}}.

\bibitem[Chaudhry et~al.(2018{\natexlab{b}})Chaudhry, Ranzato, Rohrbach, and Elhoseiny]{A-GEN}
Chaudhry, A., Ranzato, M., Rohrbach, M., and Elhoseiny, M.
\newblock Efficient lifelong learning with a-gem.
\newblock In \emph{International Conference on Learning Representations}, 2018{\natexlab{b}}.

\bibitem[Chen et~al.(2024)Chen, Wang, Yao, Bai, Hou, and Li]{saftty_neuron}
Chen, J., Wang, X., Yao, Z., Bai, Y., Hou, L., and Li, J.
\newblock Finding safety neurons in large language models.
\newblock \emph{arXiv preprint arXiv:2406.14144}, 2024.

\bibitem[Dai et~al.(2022)Dai, Dong, Hao, Sui, Chang, and Wei]{factual_info}
Dai, D., Dong, L., Hao, Y., Sui, Z., Chang, B., and Wei, F.
\newblock Knowledge neurons in pretrained transformers.
\newblock In \emph{Proceedings of the 60th Annual Meeting of the Association for Computational Linguistics (Volume 1: Long Papers)}, Jan 2022.
\newblock \doi{10.18653/v1/2022.acl-long.581}.
\newblock URL \url{http://dx.doi.org/10.18653/v1/2022.acl-long.581}.

\bibitem[Dohare et~al.(2024)Dohare, Hernandez-Garcia, Lan, Rahman, Mahmood, and Sutton]{dohare2024loss}
Dohare, S., Hernandez-Garcia, J.~F., Lan, Q., Rahman, P., Mahmood, A.~R., and Sutton, R.~S.
\newblock Loss of plasticity in deep continual learning.
\newblock \emph{Nature}, 632\penalty0 (8026):\penalty0 768--774, 2024.

\bibitem[Dravid et~al.(2023)Dravid, Gandelsman, Efros, and Shocher]{rosetta_neuron}
Dravid, A., Gandelsman, Y., Efros, A.~A., and Shocher, A.
\newblock Rosetta neurons: Mining the common units in a model zoo.
\newblock In \emph{Proceedings of the IEEE/CVF International Conference on Computer Vision}, pp.\  1934--1943, 2023.

\bibitem[eMermillod et~al.(2013)eMermillod, eBugaiska, and eBONIN]{dilemma}
eMermillod, M., eBugaiska, A., and eBONIN, P.
\newblock The stability-plasticity dilemma: Investigating the continuum from catastrophic forgetting to age-limited learning effects.
\newblock \emph{Frontiers in Psychology,Frontiers in Psychology}, Aug 2013.

\bibitem[Foundation(2024)]{atari_gymnasium}
Foundation, F.
\newblock Atari environments in gymnasium.
\newblock \url{https://gymnasium.farama.org/environments/atari/}, 2024.
\newblock URL \url{https://gymnasium.farama.org/environments/atari/}.
\newblock Accessed: 2024-09-14.

\bibitem[Goodfellow et~al.(2015)Goodfellow, Mirza, Courville, and Bengio]{stability1}
Goodfellow, I.~J., Mirza, M., Courville, A., and Bengio, Y.
\newblock An empirical investigation of catastrophic forgetting in gradient-based neural networks.
\newblock \emph{stat}, 1050:\penalty0 4, 2015.

\bibitem[Gurnee \& Tegmark(2023)Gurnee and Tegmark]{input_activate2}
Gurnee, W. and Tegmark, M.
\newblock Language models represent space and time.
\newblock In \emph{The Twelfth International Conference on Learning Representations}, Oct 2023.

\bibitem[Haarnoja et~al.(2018)Haarnoja, Zhou, Abbeel, and Levine]{sac}
Haarnoja, T., Zhou, A., Abbeel, P., and Levine, S.
\newblock Soft actor-critic: Off-policy maximum entropy deep reinforcement learning with a stochastic actor.
\newblock In \emph{International conference on machine learning}, pp.\  1861--1870. PMLR, 2018.

\bibitem[Huang et~al.(2022)Huang, Dossa, Ye, Braga, Chakraborty, Mehta, and Ara{\~A}{\v{s}}jo]{cleanrl}
Huang, S., Dossa, R. F.~J., Ye, C., Braga, J., Chakraborty, D., Mehta, K., and Ara{\~A}{\v{s}}jo, J.~G.
\newblock Cleanrl: High-quality single-file implementations of deep reinforcement learning algorithms.
\newblock \emph{Journal of Machine Learning Research}, 23\penalty0 (274):\penalty0 1--18, 2022.

\bibitem[Kim et~al.(2023)Kim, Noci, Orvieto, and Hofmann]{aux}
Kim, S., Noci, L., Orvieto, A., and Hofmann, T.
\newblock Achieving a better stability-plasticity trade-off via auxiliary networks in continual learning.
\newblock \emph{CVPR2023}, Mar 2023.

\bibitem[Kiran et~al.(2021)Kiran, Sobh, Talpaert, Mannion, Al~Sallab, Yogamani, and P{\'e}rez]{autodriveing}
Kiran, B.~R., Sobh, I., Talpaert, V., Mannion, P., Al~Sallab, A.~A., Yogamani, S., and P{\'e}rez, P.
\newblock Deep reinforcement learning for autonomous driving: A survey.
\newblock \emph{IEEE Transactions on Intelligent Transportation Systems}, 23\penalty0 (6):\penalty0 4909--4926, 2021.

\bibitem[Kirkpatrick et~al.(2017)Kirkpatrick, Pascanu, Rabinowitz, Veness, Desjardins, Rusu, Milan, Quan, Ramalho, Grabska-Barwinska, Hassabis, Clopath, Kumaran, and Hadsell]{ewc}
Kirkpatrick, J., Pascanu, R., Rabinowitz, N., Veness, J., Desjardins, G., Rusu, A.~A., Milan, K., Quan, J., Ramalho, T., Grabska-Barwinska, A., Hassabis, D., Clopath, C., Kumaran, D., and Hadsell, R.
\newblock Overcoming catastrophic forgetting in neural networks.
\newblock \emph{Proceedings of the National Academy of Sciences}, pp.\  3521–3526, Mar 2017.
\newblock \doi{10.1073/pnas.1611835114}.
\newblock URL \url{http://dx.doi.org/10.1073/pnas.1611835114}.

\bibitem[Kumar et~al.(2023)Kumar, Marklund, and Van~Roy]{regularization}
Kumar, S., Marklund, H., and Van~Roy, B.
\newblock Maintaining plasticity in continual learning via regenerative regularization.
\newblock 2023.

\bibitem[Liu et~al.(2024)Liu, Obando-Ceron, Courville, and Pan]{neuroplastic}
Liu, J., Obando-Ceron, J., Courville, A., and Pan, L.
\newblock Neuroplastic expansion in deep reinforcement learning.
\newblock \emph{arXiv preprint arXiv:2410.07994}, 2024.

\bibitem[Lopez-Paz \& Ranzato(2017)Lopez-Paz and Ranzato]{FWT}
Lopez-Paz, D. and Ranzato, M.
\newblock Gradient episodic memory for continual learning.
\newblock \emph{Advances in neural information processing systems}, 30, 2017.

\bibitem[Ma et~al.(2024)Ma, Li, Zhang, Liu, Wang, Chen, Shen, Wang, and Tao]{visual}
Ma, G., Li, L., Zhang, S., Liu, Z., Wang, Z., Chen, Y., Shen, L., Wang, X., and Tao, D.
\newblock Revisiting plasticity in visual reinforcement learning: Data, modules and training stages.
\newblock In \emph{The Twelfth International Conference on Learning Representations}, 2024.

\bibitem[Mallya \& Lazebnik(2018)Mallya and Lazebnik]{Packnet}
Mallya, A. and Lazebnik, S.
\newblock Packnet: Adding multiple tasks to a single network by iterative pruning.
\newblock In \emph{Proceedings of the IEEE conference on Computer Vision and Pattern Recognition}, pp.\  7765--7773, 2018.

\bibitem[McCloskey \& Cohen(1989)McCloskey and Cohen]{stability}
McCloskey, M. and Cohen, N.~J.
\newblock Catastrophic interference in connectionist networks: The sequential learning problem.
\newblock In \emph{Psychology of learning and motivation}, volume~24, pp.\  109--165. Elsevier, 1989.

\bibitem[Mendez et~al.(2022)Mendez, van Seijen, and Eaton]{NC}
Mendez, J.~A., van Seijen, H., and Eaton, E.
\newblock Modular lifelong reinforcement learning via neural composition.
\newblock \emph{arXiv preprint arXiv:2207.00429}, 2022.

\bibitem[Mnih et~al.(2013)Mnih, Kavukcuoglu, Silver, Graves, Antonoglou, Wierstra, and Riedmiller]{atari}
Mnih, V., Kavukcuoglu, K., Silver, D., Graves, A., Antonoglou, I., Wierstra, D., and Riedmiller, M.
\newblock Playing atari with deep reinforcement learning.
\newblock \emph{arXiv preprint arXiv:1312.5602}, 2013.

\bibitem[Mnih et~al.(2015)Mnih, Kavukcuoglu, Silver, Rusu, Veness, Bellemare, Graves, Riedmiller, Fidjeland, Ostrovski, et~al.]{score}
Mnih, V., Kavukcuoglu, K., Silver, D., Rusu, A.~A., Veness, J., Bellemare, M.~G., Graves, A., Riedmiller, M., Fidjeland, A.~K., Ostrovski, G., et~al.
\newblock Human-level control through deep reinforcement learning.
\newblock \emph{nature}, 518\penalty0 (7540):\penalty0 529--533, 2015.

\bibitem[Molchanov et~al.(2022)Molchanov, Tyree, Karras, Aila, and Kautz]{expression}
Molchanov, P., Tyree, S., Karras, T., Aila, T., and Kautz, J.
\newblock Pruning convolutional neural networks for resource efficient inference.
\newblock In \emph{International Conference on Learning Representations}, 2022.

\bibitem[Nikishin et~al.(2022{\natexlab{a}})Nikishin, Schwarzer, D’Oro, Bacon, and Courville]{plasticity1}
Nikishin, E., Schwarzer, M., D’Oro, P., Bacon, P.-L., and Courville, A.
\newblock The primacy bias in deep reinforcement learning.
\newblock In \emph{International conference on machine learning}, pp.\  16828--16847. PMLR, 2022{\natexlab{a}}.

\bibitem[Nikishin et~al.(2022{\natexlab{b}})Nikishin, Schwarzer, D’Oro, Bacon, and Courville]{primacy_bias}
Nikishin, E., Schwarzer, M., D’Oro, P., Bacon, P.-L., and Courville, A.
\newblock The primacy bias in deep reinforcement learning.
\newblock In \emph{International conference on machine learning}, pp.\  16828--16847. PMLR, 2022{\natexlab{b}}.

\bibitem[Nikishin et~al.(2024)Nikishin, Oh, Ostrovski, Lyle, Pascanu, Dabney, and Barreto]{plasticity_injection}
Nikishin, E., Oh, J., Ostrovski, G., Lyle, C., Pascanu, R., Dabney, W., and Barreto, A.
\newblock Deep reinforcement learning with plasticity injection.
\newblock \emph{Advances in Neural Information Processing Systems}, 36, 2024.

\bibitem[Paik et~al.(2019)Paik, Oh, Kwak, and Kim]{important_neuron}
Paik, I., Oh, S., Kwak, T.-Y., and Kim, I.
\newblock Overcoming catastrophic forgetting by neuron-level plasticity control.
\newblock \emph{AAAI2020}, Jul 2019.

\bibitem[Sajjad et~al.(2022)Sajjad, Durrani, and Dalvi]{neuron}
Sajjad, H., Durrani, N., and Dalvi, F.
\newblock Neuron-level interpretation of deep nlp models: A survey.
\newblock \emph{Transactions of the Association for Computational Linguistics}, 10:\penalty0 1285--1303, 2022.

\bibitem[Sokar et~al.(2023)Sokar, Agarwal, Castro, and Evci]{dormant_neuron}
Sokar, G., Agarwal, R., Castro, P.~S., and Evci, U.
\newblock The dormant neuron phenomenon in deep reinforcement learning.
\newblock In \emph{International Conference on Machine Learning}, pp.\  32145--32168. PMLR, 2023.

\bibitem[Sutton(2018)]{sutton}
Sutton, R.~S.
\newblock Reinforcement learning: An introduction.
\newblock \emph{A Bradford Book}, 2018.

\bibitem[Tang et~al.(2024)Tang, Luo, Huang, Zhang, Wang, Zhao, Wei, and Wen]{Language-specific_neurons}
Tang, T., Luo, W., Huang, H., Zhang, D., Wang, X., Zhao, X., Wei, F., and Wen, J.-R.
\newblock Language-specific neurons: The key to multilingual capabilities in large language models.
\newblock \emph{arXiv preprint arXiv:2402.16438}, 2024.

\bibitem[Wang et~al.(2024)Wang, Zhang, Su, and Zhu]{AIA}
Wang, L., Zhang, X., Su, H., and Zhu, J.
\newblock A comprehensive survey of continual learning: theory, method and application.
\newblock \emph{IEEE Transactions on Pattern Analysis and Machine Intelligence}, 2024.

\bibitem[Wang et~al.(2022)Wang, Wen, Zhang, Hou, Liu, and Li]{skill_neuron}
Wang, X., Wen, K., Zhang, Z., Hou, L., Liu, Z., and Li, J.
\newblock Finding skill neurons in pre-trained transformer-based language models.
\newblock In \emph{Proceedings of the 2022 Conference on Empirical Methods in Natural Language Processing}, pp.\  11132--11152, 2022.

\bibitem[Wolczyk et~al.(2022)Wolczyk, Zaj{{a}}c, Pascanu, Kuci{\'n}ski, and Mi{\l}o{\'s}]{clone-sac}
Wolczyk, M., Zaj{{a}}c, M., Pascanu, R., Kuci{\'n}ski, {\L}., and Mi{\l}o{\'s}, P.
\newblock Disentangling transfer in continual reinforcement learning.
\newblock \emph{Advances in Neural Information Processing Systems}, 35:\penalty0 6304--6317, 2022.

\bibitem[Xu et~al.(2024)Xu, Zhan, Wong, and Chao]{roles}
Xu, H., Zhan, R., Wong, D.~F., and Chao, L.~S.
\newblock Let's focus on neuron: Neuron-level supervised fine-tuning for large language model.
\newblock \emph{arXiv preprint arXiv:2403.11621}, 2024.

\bibitem[Yang et~al.(2020)Yang, Xu, Wu, and Wang]{SM}
Yang, R., Xu, H., Wu, Y., and Wang, X.
\newblock Multi-task reinforcement learning with soft modularization.
\newblock \emph{Advances in Neural Information Processing Systems}, 33:\penalty0 4767--4777, 2020.

\bibitem[Yang et~al.(2023)Yang, Zhou, Jiang, Long, and Shi]{CoTASP}
Yang, Y., Zhou, T., Jiang, J., Long, G., and Shi, Y.
\newblock Continual task allocation in meta-policy network via sparse prompting.
\newblock In \emph{International Conference on Machine Learning}, pp.\  39623--39638. PMLR, 2023.

\bibitem[Yu et~al.(2020)Yu, Quillen, He, Julian, Hausman, Finn, and Levine]{meta-world}
Yu, T., Quillen, D., He, Z., Julian, R., Hausman, K., Finn, C., and Levine, S.
\newblock Meta-world: A benchmark and evaluation for multi-task and meta reinforcement learning.
\newblock In \emph{Conference on robot learning}, pp.\  1094--1100. PMLR, 2020.

\end{thebibliography}
\bibliographystyle{icml2025}

%%%%%%%%%%%%%%%%%%%%%%%%%%%%%%%%%%%%%%%%%%%%%%%%%%%%%%%%%%%%%%%%%%%%%%%%%%%%%%%
%%%%%%%%%%%%%%%%%%%%%%%%%%%%%%%%%%%%%%%%%%%%%%%%%%%%%%%%%%%%%%%%%%%%%%%%%%%%%%%
% APPENDIX
%%%%%%%%%%%%%%%%%%%%%%%%%%%%%%%%%%%%%%%%%%%%%%%%%%%%%%%%%%%%%%%%%%%%%%%%%%%%%%%
%%%%%%%%%%%%%%%%%%%%%%%%%%%%%%%%%%%%%%%%%%%%%%%%%%%%%%%%%%%%%%%%%%%%%%%%%%%%%%%
\newpage
\appendix
\onecolumn
\section{Related Wrok}
\textbf{Balance between stability and plasticity}.
In DRL, the agent faces a fundamental challenge: the stability-plasticity dilemma, first introduced by \citet{Carpenter_Grossberg}. Recent research has proposed various strategies to address this issue by balancing stability and plasticity.

Replay-based methods are widely employed to enhance stability by reusing experiences from past distributions. For example, \citet{A-GEN} introduced A-GEM, which combines episodic memory to ensure that the average loss of prior tasks does not increase when learning a new task. Similarly, \citet{clone-sac} proposed ClonEx-SAC, which uses actor behavioral cloning and best-return exploration to boost performance in CRL. To reduce storage requirements, pseudo-rehearsals generated from a generative model have also been proposed \citep{Pseudo-rehearsal}.

Maintaining the expressiveness of neurons is key to preserving plasticity. \citet{primacy_bias} proposed a mechanism that periodically resets a portion of the agent's network to counteract plasticity loss. Likewise, \citet{plasticity_injection} introduced plasticity injection, a lightweight intervention that enhances network plasticity without increasing trainable parameters or introducing prediction bias. The Reset \& Distillation (R\&D) framework combines resetting the online actor-critic network for new tasks with offline distillation of knowledge from previous action probabilities, effectively retaining plasticity \citep{replay}. Additionally, \citet{plasticity2} proposed the Concatenated ReLUs (CReLUs) activation function to prevent activation collapse, thereby alleviating plasticity degradation.

Modularity-based approaches have shown promise in balancing stability and plasticity by decoupling task-specific and general knowledge. For instance, \citet{modularity1} decomposed the value function into a permanent value function, which captures persistent knowledge, and a transient value function, which facilitates rapid adaptation. \citet{SM} designed a routing network to estimate task-specific routing strategies, reconfigure the base network, and combine routes using a soft modularity mechanism, making it effective for sequential tasks. Similarly, \citet{NC} proposed a compositional lifelong RL framework that uses accumulated neural components to accelerate learning for new tasks while preserving performance on past tasks via offline RL and replayed experiences.

\textbf{Neuron-level Research}
Recent research highlights that not all neurons remain active across varying contexts, and this neuron sparsity is often positively correlated with task-specific performance \citep{roles}. Building on this insight, numerous studies have focused on identifying and leveraging skill neurons to interpret network behavior and tackle specific challenges, achieving significant advancements. For example, skill neurons in pre-trained Transformers, which demonstrate strong predictive value for task labels, have been utilized for network pruning to enhance efficiency and improve transferability \citep{skill_neuron}. \citet{dormant_neuron} investigate dormant neurons in deep reinforcement learning and propose a method to recycle them during training. Similarly, \citet{rosetta_neuron} introduce Rosetta Neurons, enabling cross-class alignments and transformations without specialized training. In large language models, language-specific neurons have been identified to control output languages by selective activation or deactivation \citep{Language-specific_neurons}, while safety neurons have been analyzed to enhance safety alignment through mechanistic interpretability \citep{saftty_neuron}.

Despite these achievements, the exploration of skill neurons in DRL remains limited. Existing neuron-level approaches primarily focus on task-specific sub-network selection. For instance, CoTASP learns hierarchical dictionaries and meta-policies to generate sparse prompts and extract sub-networks as task-specific policies \citep{CoTASP}. Similarly, \citet{Packnet} sequentially allocate multiple tasks within a single network through iterative pruning and re-training, balancing performance and storage efficiency. Unlike these methods, our work identifies RL skill neurons specifically tailored to deep reinforcement learning, ensuring a balance between stability and plasticity by preserving the task-relevant knowledge encoded in these neurons while allowing for fine-tuning.

\section{Preliminary}
\subsection{Markov Decision Process (MDP)}
A Markov Decision Process(MDP) is a framework used to describe a problem involving learning from actions to achieve a goal. Almost all reinforcement learning problems can be characterized as a Markov Decision Process. Each MDP is defined by a tuple $<S, A, P, R, \gamma >$, where $S$ and $A$ represent state and action spaces respectively. The transition dynamics of the MDP are defined by the function $P: S\times A\times S \rightarrow [0,1]$, which represents the probability of transitioning from a give state $s$ with action $a$ to state $s^{\prime}$. The reward function is represented by $R: S\times A\times S\rightarrow \mathbb{R}$, and $\gamma \in (0,1)$ is the discount factor. At each time step $t$, an agent observes the state of the environment, denoted as $s_t$, and selects an action $a_t$ according to a policy $\pi(a|s)$. One time step later, the agent receives a numerical reward $r_{t+1}$ and transitions to a new state $s_{t+1}$. In the simplest case, the return is the sum of the rewards when the agent–environment interaction naturally breaks into subsequences, which we refer to episodes \citep{sutton}.

\subsection{Soft Actor-Critic (SAC)}
Soft Actor-Critic (SAC) is an off-policy actor-critic deep reinforcement learning algorithm that leverages maximum entropy to promote exploration. This work employs SAC to train a policy that effectively balances stability and plasticity , chosen for its sample efficiency, excellent performance, and robust stability. In this framework, the actor aims to maximize both the expected reward and the entropy of the policy. The parameters $\phi$ of the actor are optimized by minimizing the following loss function:
$$
J_\pi(\phi)=E_{s_t\sim D , a_t\sim \pi_\phi} [\alpha log\pi_\phi(a_t | s_t) - Q_\theta(s_t,a_t)],
$$

where $D$ is the replay buffer, $\alpha $ is the temperature parameter controlling the trade-off between exploration and exploitation, $\theta $ denotes the parameters of the critic network, $\pi_\phi$ represents the policy learned by the actor $\phi$ , and $Q_\theta$ denotes the Q-value estimated by the critic $\theta$.  The critic network is trained to minimize the squared residual error:
$$
J_Q(\theta)=E_{(s_t,a_t,s_{t+1})\sim D} [\frac{1}{2}(Q_\theta (s_t,a_t)-r_t-\gamma \hat{V}(s_{t+1})],
$$
$$
\hat{V}(s_t)=E_{a_t\sim \pi_\phi } [Q_\theta  (s_t,a_t )-\alpha log\pi_\phi  (a_t | s_t )],
$$
where $\gamma$ represents the discount factor.

\subsection{Neuron}
In neural networks, various components, such as blocks and layers, play distinct roles. Here, we define a neuron as a single output dimension from a layer. For example, in a fully connected layer, each output dimension corresponds to a neuron. Similarly, in a convolutional layer, each output channel represents a neuron. Furthermore, following the terminology used by \citet{neuron}, we classify neurons that encapsulate a single concept as focused neurons, while a group of neurons collectively representing a concept are termed group neurons.

\section{Experiment}

\subsection{Baseline}
\label{baseline}
\textbf{EWC}: Elastic Weight Consolidation (EWC)~\citep{ewc} addresses the challenge of catastrophic forgetting by allowing neural networks to retain proficiency in previously learned tasks even after a long hiatus. It achieves this by selectively slowing down learning for weights that are crucial for retaining knowledge of these tasks. This approach has demonstrated excellent performance in sequentially solving a series of classification tasks, such as those in the MNIST handwritten digit dataset, and in learning several Atari 2600 games sequentially.

\textbf{NPC}: Neuron-level Plasticity Control (NPC)~\citep{important_neuron} preserves the existing knowledge from the previous tasks by controlling the plasticity of the network at the neuron level. NPC estimates the importance value of each neuron and consolidates important neurons by applying lower learning rates, rather than restricting individual connection weights to stay close to the values optimized for the previous tasks. The experimental results on the several classification datasets show that neuron-level consolidation is substantially effective.

\textbf{ANCL}: Auxiliary Network Continual Learning (ANCL) is an innovative approach that incorporates an auxiliary network to enhance plasticity within a model that primarily emphasizes stability. Specifically, this framework introduces a regularizer that effectively balances plasticity and stability, achieving superior performance over strong baselines in both task-incremental and class-incremental learning scenarios.

\textbf{CoTASP}: Continual Task Allocation via Sparse Prompting (CoTASP)~\citep{CoTASP} learns over-complete dictionaries to produce sparse masks as prompts extracting a sub-network for each task from a meta-policy network. Hence, relevant tasks share more neurons in the meta-policy network due to similar prompts while cross-task interference causing forgetting is effectively restrained. It outperforms existing continual and multi-task RL methods on all seen tasks, forgetting reduction, and generalization to unseen tasks.

\textbf{CRelu}: Concatenated ReLUs (CReLUs)~\citep{plasticity2} is a simple activation function that concatenates the input with its negation and applies ReLU to the result. It performs effectively in facilitating continual learning in a changing environment.

\textbf{CBP}: Continual BackPropagation (CBP)~\citep{dohare2024loss} reinitializes a small number of units during training, typically fewer than one per step. To prevent disruption of what the network has already learned, only the least-used units are considered for reinitialization. It shows great performance on Continual ImageNet and class-incremental CIFAR-100.

\textbf{PI}: Plasticity Injection (PI)~\citep{plasticity_injection} freeze the parameters $\theta$ and introduce a new set of parameters $\theta \prime$ sampled from random initialization at some point in training, where the network might have started losing plasticity. The results on Atari show that plasticity injection attains stronger performance compared to alternative methods while being computationally efficient.

\subsection{Benchmark}
\label{benchmark}

\textbf{Meta-World}. Meta-World is an open-source benchmark for meta-reinforcement learning and multitask learning, comprising 50 distinct robotic manipulation tasks \citep{meta-world}.

All tasks are executed by a simulated Sawyer robot, with the action space defined as a 2-tuple: the change in the 3D position of the end-effector, followed by a normalized torque applied to the gripper fingers.

The observation space has a consistent dimensionality of 39, although different dimensions correspond to various aspects of each task. Typically, the observation space is represented as a 6-tuple, including the 3D Cartesian position of the end-effector, a normalized measure of the gripper’s openness, the 3D position and the quaternion of the first object, the 3D position and quaternion of the second object, all previous measurements within the environment, and the 3D position of the goal.

\begin{figure}[htbp]
% \centering
\begin{center}
%\framebox[4.0]{$\;$}
\includegraphics[width=\columnwidth]{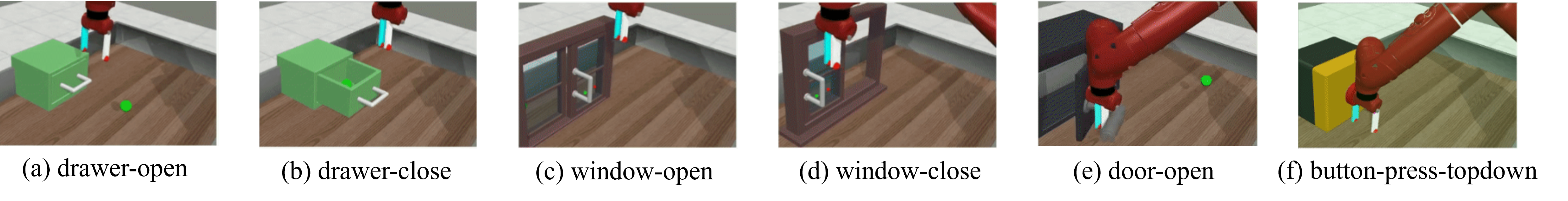}
\end{center}
\caption{Tasks in the Meta-World benchmark used in our experiments.}
\label{fig:meta-world_task}
\end{figure}

The reward function for all tasks is structured and multi-component, aiding in effective policy learning for each task component. With this design, the reward functions maintain a similar magnitudes across tasks, generally ranging between 0 and 10. The descriptions of the six tasks used in our experiments are listed below, and the appearance of these tasks is shown in Figure \ref{fig:meta-world_task}.

\begin{itemize}
    \item \textbf{drawer-open}:
        Open a drawer, with randomized drawer positions.
    \item \textbf{drawer-close}:
        Push and close a drawer, with randomized drawer positions.
    \item \textbf{window-open}:
        Push and open a window, with randomized window positions.
    \item \textbf{window-close}:
        Push and close a window, with randomized window positions.
    \item \textbf{door-open}:
        Open a door with a revolving joint. Randomize door positions.
    \item \textbf{button-press-topdown}:
        Press a button from the top. Randomize button positions.
\end{itemize}

\textbf{Atari}.
Atari environments are simulated using the Arcade Learning Environment (ALE) \citep{ale} via the Stella emulator.

Each environment utilizes a subset of the full action space, which includes actions like NOOP, FIRE, UP, RIGHT, LEFT, DOWN, UPRIGHT, UPLEFT, DOWNRIGHT, DOWNLEFT, UPFIRE, RIGHTFIRE, LEFTFIRE, DOWNFIRE, UPRIGHTFIRE, UPLEFTFIRE, DOWNRIGHTFIRE, and DOWNLEFTFIRE. By default, most environments employ only a smaller subset of these actions, excluding those that have no effect on gameplay.

Observations in Atari environments are RGB images displayed to human players, with $obs\_type="rgb"$, corresponding to an observation space defined as $Box(0, 255, (210, 160, 3), np.uint8)$.

The specific reward dynamics vary depending on the environment and are typically detailed in the game’s manual.

\begin{figure}[htbp]
% \centering
\begin{center}
%\framebox[4.0]{$\;$}
\includegraphics[width=0.6\columnwidth]{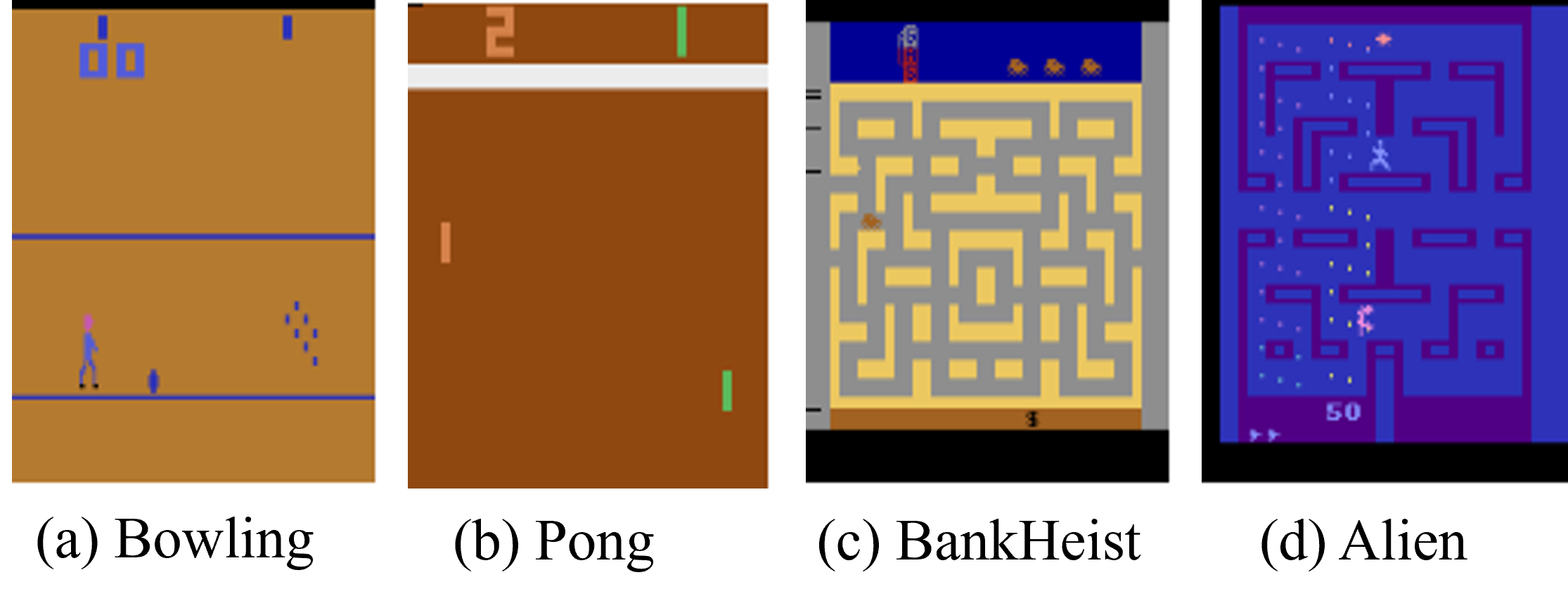}
\end{center}
\caption{Games in the Atari benchmark used in our experiments.}
\label{fig:atari_game}
\end{figure}

The descriptions of the four games used in our experiments are listed below \citep{atari_gymnasium}, and the appearance of these games is shown in Figure \ref{fig:atari_game}.
\begin{itemize}
    \item \textbf{Bowling}:
        The goal is to score as many points as possible in a 10-frame game. Each frame allows up to two tries. Knocking down all pins on the first try is called a "strike", while doing so on the second try is a "spare". Failing to knock down all pins in two attempts results in an "open" frame.
    \item \textbf{Pong}:
        You control the right paddle and compete against the computer-controlled left paddle. The objective is to deflect the ball away from your goal and into the opponent's goal.
    \item \textbf{BankHeist}:
        You play as a bank robber trying to rob as many banks as possible while avoiding the police in maze-like cities. You can destroy police cars using dynamite and refill your gas tank by entering new cities. Lives are lost if you run out of gas, are caught by the police, or run over your own dynamite.
    \item \textbf{Alien}:
         You are trapped in a maze-like spaceship with three aliens. Your goal is to destroy their eggs scattered throughout the ship while avoiding the aliens. You have a flamethrower to fend them off and can occasionally collect a power-up (pulsar) that temporarily enables you to kill aliens.
\end{itemize}

\subsection{Experiment setting}
\label{setting}
For all experiments, we utilize the open-source PyTorch implementation of Soft Actor-Critic (SAC) provided by CleanRL \citep{cleanrl} on a single RTX2080Ti GPU. CleanRL is a Deep Reinforcement Learning library that offers high-quality, single-file implementations with research-friendly features. The code is both clean and straightforward, and we adhere to the configurations provided by CleanRL.
During training, we employ an $\epsilon$-greedy exploration policy at the start, setting $\epsilon = 1$ for the first $10^4$ time steps to promote exploration. The environment is wrapped using Gym wrappers to facilitate experimentation. For the Meta-World benchmark, we utilize the RecordEpisodeStatistics wrapper to gather episode statistics. For the Atari benchmark, in addition to RecordEpisodeStatistics, we preprocess the $210\times160$ pixel images by downsampling them to $84\times84$ using bilinear interpolation, converting the RGB images to the YUV format, and using only the grayscale channel. Additionally, we set a maximum limit on the number of noop and skip steps to standardize the exploration.

Regarding network architecture, we use the same actor and critic networks for all tasks within the same benchmark to ensure consistency. For the Meta-World benchmark, we employ a neural network comprising four fully connected layers, of which the hidden size is [768, 768, 768]. For the Atari benchmark, we use a convolutional neural network (CNN) with three convolutional layers featuring 32, 64, and 64 channels, respectively, followed by three fully connected layers, of which the hidden size is [768, 768].

To reduce randomness and enhance the reliability of our results, we train each agent using three random seeds. Additional hyper-parameters for the SAC algorithm applied in the Meta-World and Atari benchmarks are detailed in Table \ref{tab:hyper-parameters}.

\begin{table}[htbp]
\caption{Hyper-parameters of SAC in our experiments.}
\label{tab:hyper-parameters}
\centering
\vspace{1em}
\begin{tabular}{ccc}
\hline
\textbf{Parameters}                          & \textbf{Values for Meta-World} & \multicolumn{1}{l}{\textbf{Values for Atari}} \\ \hline
Initial collect steps                       & 10000                         & 20000                                        \\
Discount factor                             & 0.99                          & 0.99                                         \\
Training environment steps                  & $10^6$                        & $1.5 \times 10^6$, $3 \times 10^6$                            \\
Testing environment steps                   & $10^5$                        & $10^5$                           \\
Replay buffer size                          & $10^6$                        & $2 \times 10^5$                              \\
Updates per environment step (Replay Ratio) & 2                             & 4                                            \\
Target network update period                & 1                             & 8000                                         \\ 
Target smoothing coefficient                & 0.005                         & 1                                            \\
Optimizer                                   & Adam                          & Adam                                         \\
Policy learning rate                        & $3\times 10^{-4}$             & $10^{-4}$                                    \\
Q-value learning rate                      & $10^{-3}$                      & $10^{-4}$                                    \\
Minibatch size                              & 256                           & 64                                           \\
Alpha                                       & 0.2                           & 0.2                                          \\
Autotune                                    & True                           & True                                        \\
Average environment steps  of success rate   & 10                            &  -                                        \\
Stable threshold to finish  training        & 0.9                           & -                                         \\
Replay interval                             & 10                            & 10                                        \\
No-op max                                   & -                             & 30                                           \\
Target entropy scale                        & -                             & 0.89                                         \\
Storing experience size                        & $10^5$                             & $10^5$                                         \\
\hline                                                                   
\end{tabular}
\end{table}

\subsection{Metrics}
\label{metric}
For the Meta-World benchmark, the average success rate is computed over 20 episodes.
For the Atari benchmark, the success rate is replaced by the return of each episode. We normalize the return for each game to obtain summary statistics across games, as follows:
\begin{equation}
R = \frac{r_{agent}-r_{random}}{r_{human}-r_{random}},
\end{equation}
where $r_{agent}$ represents the average return evaluated over $10^5$ steps, the random score $r_{random}$ and human score $r_{human}$ are consistent with those used by \citet{score}, as detailed in Table \ref{tab:scores}.

\begin{table}[htbp]
\caption{Normalization scores of Atari games.}
\label{tab:scores}
\centering
\vspace{1em}
\begin{tabular}{ccc}
\hline
\textbf{games} & \textbf{$r_{random}$} & \textbf{$r_{human}$} \\ \hline
Bowling        & 23.1                 & 154.8                    \\ 
Pong           & -20.7                & 9.3                    \\ 
BankHeist      & 14.2                 & 734.4                    \\ 
Alien          & 227.5                & 6875                    \\ 
\hline
\end{tabular}%
\end{table}

For the Atari benchmark tasks, the overall performance is evaluated by Average Return (AR), which is analogous to ASR in the Meta-World benchmark. It is calculated as follows:
\begin{equation}
    AR = \frac{1}{k}\sum_{i=1}^{k}{\frac{1}{i}\sum_{i \ge j}{R_{i,j}}},
\end{equation}
where $R_{i,j}$ represents the average return evaluated on the $j$-th task after completing the learning of the $i$-th task ($i \ge j$), and $k$ represents the number of tasks. A higher AR indicates better performance in balancing stability and plasticity.

\subsection{Results on the Meta-world benchmark}
\label{result_meta-world}
\begin{figure}[htbp]
\centering
\begin{center}
%\framebox[4.0in]{$\;$}
\vspace{0.5em}
\includegraphics[width=\linewidth]{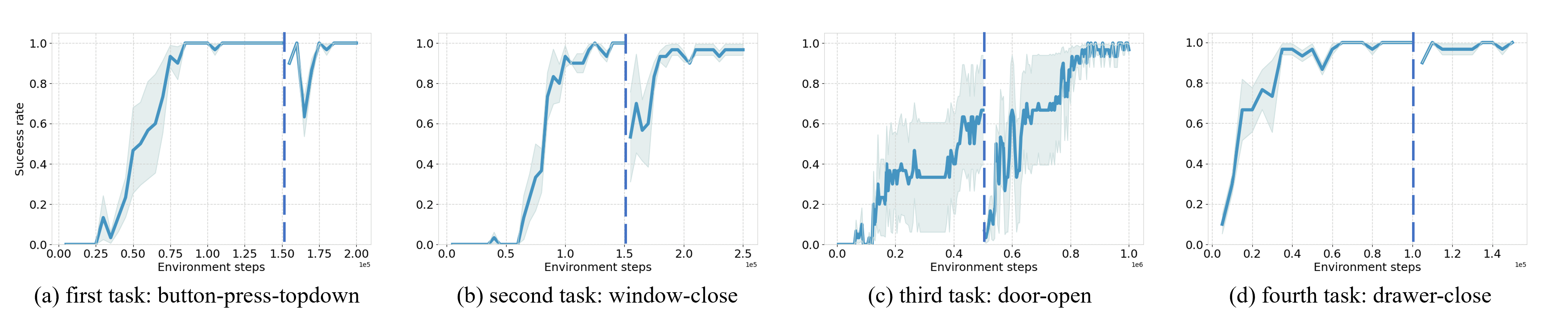}
\end{center}
\caption{Training process of NBSP on (button-press-topdown $\rightarrow$ window-close $\rightarrow$ door-open $\rightarrow$ drawer-close) cycling task.}
\label{fig:training_process_4_task}
\end{figure}

\begin{figure}[htbp]
\centering
\begin{center}
%\framebox[4.0in]{$\;$}
\vspace{0.5em}
\includegraphics[width=0.7\linewidth]{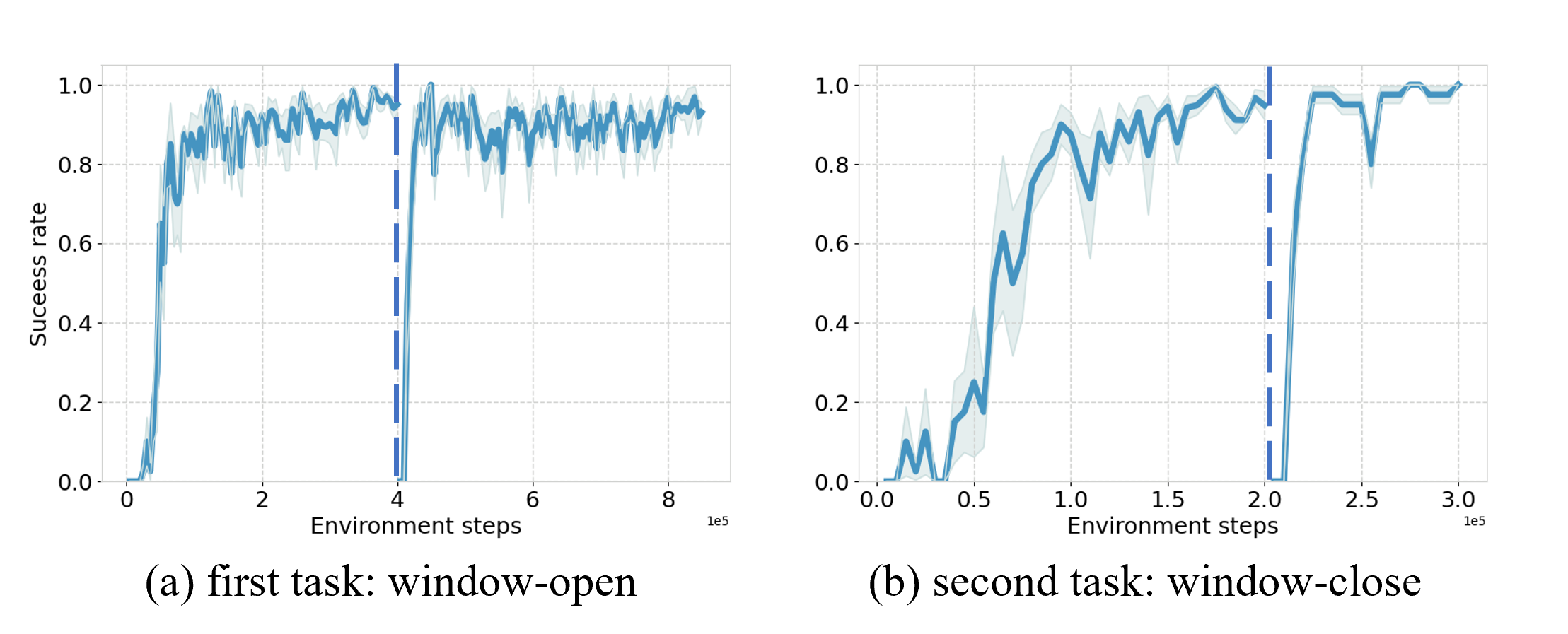}
\end{center}
\caption{Training process of NBSP on (window-open $\rightarrow$ window-close) cycling task.}
\label{fig:training_process_window-open-close}
\end{figure}

\begin{figure}[htbp]
\centering
\begin{center}
%\framebox[4.0in]{$\;$}
\vspace{0.5em}
\includegraphics[width=0.7\linewidth]{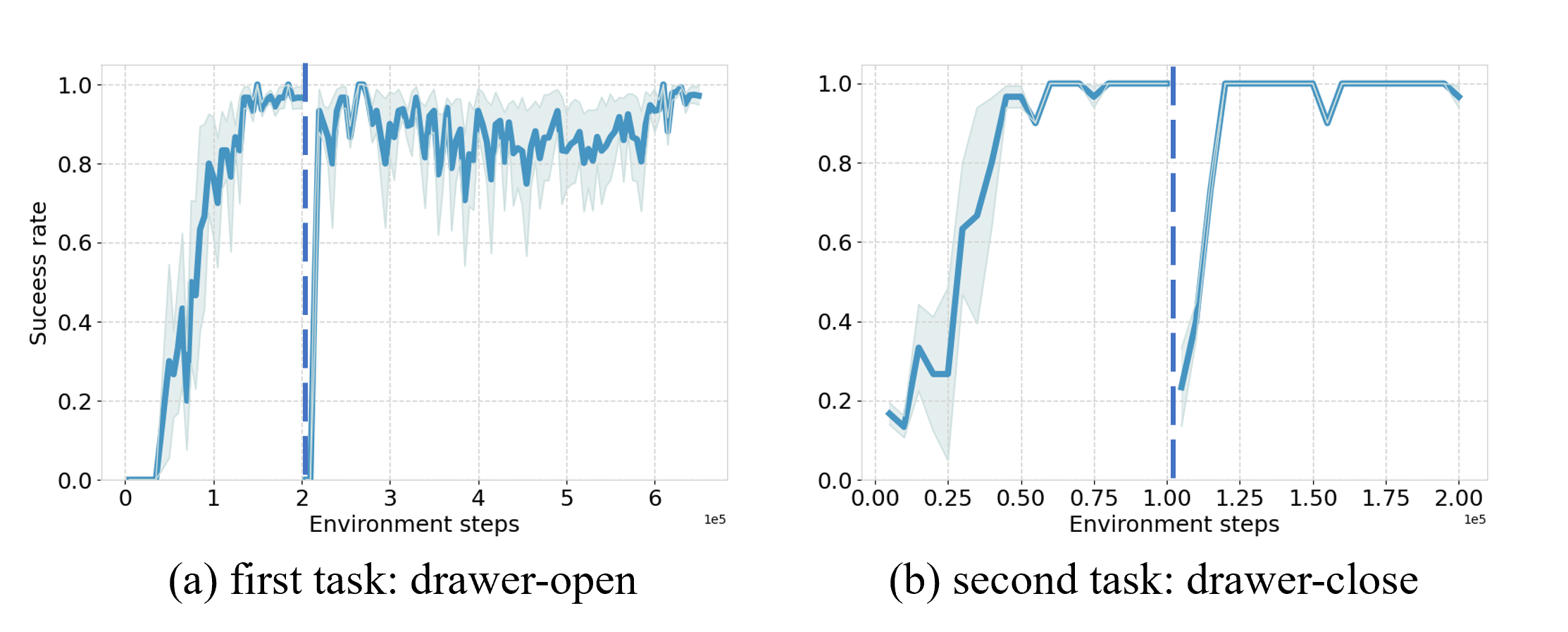}
\end{center}
\caption{Training process of NBSP on (drawer-open $\rightarrow$ drawer-close) cycling task.}
\label{fig:training_process_drawer-open-close}
\end{figure}

\begin{figure}[htbp]
\centering
\begin{center}
%\framebox[4.0in]{$\;$}
\vspace{0.5em}
\includegraphics[width=0.7\linewidth]{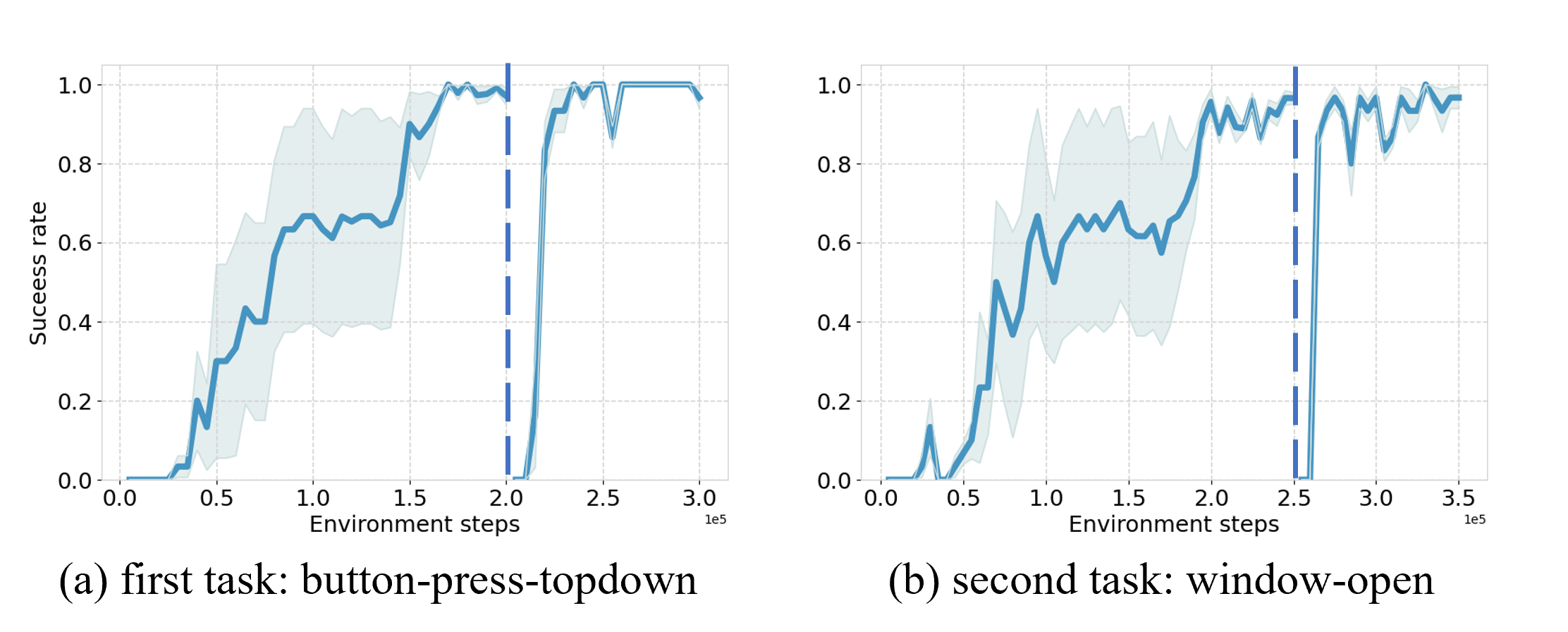}
\end{center}
\caption{Training process of NBSP on (button-press-topdown $\rightarrow$ window-open) cycling task.}
\label{fig:training_process_button-window-open}
\end{figure}

The training process of the other four-tasks cycling task is shown in Figure \ref{fig:training_process_4_task}, and those of the two-task cycling tasks are shown in Figure \ref{fig:training_process_window-open-close}, Figure \ref{fig:training_process_drawer-open-close} and Figure \ref{fig:training_process_button-window-open} respectively. The same as found in Section \ref{experiment_meta-world}, during the second cycle of learning the same task, the agent is able to master the task more rapidly. 

\subsection{Ablation study}
\label{ablation_component}

The results of the ablation study on two critical components, gradient masking and experience replay techniques, are shown in Table \ref{tab:components_metrics_window} for the (window-open $\rightarrow$ window-close) cycling task and in Table \ref{tab:components_metrics_drawer} for the (drawer-open $\rightarrow$ drawer-close) cycling task. From these results, it is evident that both gradient masking and experience replay techniques independently contribute to improving the stability of the agent while maintain great plasticity. Furthermore, combining both techniques yields superior performance, demonstrating the enhanced effectiveness of their integration.

\begin{table}[htbp]
\centering
\renewcommand{\arraystretch}{1} % 调整行高
\caption{Results of ablation study of gradient masking and experience replay techniques on (window-open $\rightarrow$ window-close) cycling task.}
\label{tab:components_metrics_window}
\vspace{1em}
\resizebox{0.6\columnwidth}{!}{%
\begin{tabular}{>{\centering\arraybackslash}m{1.8cm} >{\centering\arraybackslash}m{1.8cm}ccc}
\toprule
\multicolumn{2}{c}{\textbf{Component}} & \multicolumn{3}{c}{\textbf{Metrics}} \\ 
\cmidrule(lr){1-2} \cmidrule(lr){3-5}
\textbf{Gradient Masking} & \textbf{Experience Replay} & \textbf{ASR} & \textbf{FM} & \textbf{FT} \\ 
\midrule
$\times$   & $\times$   & 0.63 $\pm$ 0.02   & 0.91 $\pm$ 0.10   & 0.97 $\pm$ 0.02 \\ 
$\times$   & \checkmark & 0.81 $\pm$ 0.08   & 0.41 $\pm$ 0.13   & 0.96 $\pm$ 0.01 \\ 
\checkmark & $\times$   & 0.78 $\pm$ 0.11   & 0.54 $\pm$ 0.26   & 0.98 $\pm$ 0.01\\ 
\checkmark & \checkmark & \textbf{ 0.90 $\pm$ 0.04 }   & \textbf{ 0.18 $\pm$ 0.01 }  & \textbf{ 0.96 $\pm$ 0.02 } \\ 
\bottomrule
\end{tabular}%
}
\end{table}

\begin{table}[htbp]
\centering
\renewcommand{\arraystretch}{1} % 调整行高
\caption{Results of ablation study of gradient masking and experience replay techniques on (drawer-open $\rightarrow$ drawer-close) cycling task.}
\label{tab:components_metrics_drawer}
\vspace{1em}
\resizebox{0.6\columnwidth}{!}{%
\begin{tabular}{>{\centering\arraybackslash}m{1.8cm} >{\centering\arraybackslash}m{1.8cm}ccc}
\toprule
\multicolumn{2}{c}{\textbf{Component}} & \multicolumn{3}{c}{\textbf{Metrics}} \\ 
\cmidrule(lr){1-2} \cmidrule(lr){3-5}
\textbf{Gradient Masking} & \textbf{Experience Replay} & \textbf{ASR} & \textbf{FM} & \textbf{FT} \\ 
\midrule
$\times$   & $\times$   & 0.67 $\pm$ 0.05   & 0.78 $\pm$ 0.10   & 0.94 $\pm$ 0.04 \\ 
$\times$   & \checkmark & 0.78 $\pm$ 0.04   & 0.48 $\pm$ 0.10   & 0.97 $\pm$ 0.01 \\ 
\checkmark & $\times$   & 0.74 $\pm$ 0.01   & 0.64 $\pm$ 0.01   & 0.98 $\pm$ 0.02\\ 
\checkmark & \checkmark & \textbf{ 0.96 $\pm$ 0.02 }  & \textbf{ 0.07 $\pm$ 0.06 }  & \textbf{ 0.98 $\pm$ 0.01 } \\ 
\bottomrule
\end{tabular}%
}
\end{table}

\section{algorithm}
\label{algorithm}
The pseudo-code of the goal-oriented method to find RL skill neurons is presented in Algorithm \ref{alg:algorithm_skill-neuron}. And the pseudo-code for SAC with NBSP is presented in Algorithm \ref{alg:algorithm_nbsp}. Key differences from standard SAC are highlighted in blue. In addition to the extra input, two main modifications include the sampling process and the network update process.

\vspace{1em}
\begin{algorithm}[t]
\caption{Procedure for Identifying RL Skill Neurons}
\label{alg:algorithm_skill-neuron}
\begin{algorithmic}[1] %[1] enables line numbers
\For{each step $t$}
    \State Compute activation $a(t)$ and GPM $q(t)$ via model forward
    \State Accumulate $a(t)$ and $q(t)$ across steps
\EndFor
\State Compute the standard activation $\overline{a}$ and standard GPM $\overline{q}$ using Eq.~\ref{eq:standard_activation} and Eq.~\ref{eq:standard_GPM}
\For{each step $t$}
    \State Compute activation $a(t)$ and GPM $q(t)$ via model forward pass
    \State Compare $a(t)$ and $q(t)$ against their respective standards, $\overline{a}$ and $\overline{q}$, and record results
\EndFor
\State Compute positive accuracy $Acc$ using Eq.~\ref{eq:acc}
\State Derive scores $Score$ for each neuron using Eq.~\ref{eq:score}
\State Rank neurons based on their scores and select the top-performing neurons as RL skill neurons $\{\mathcal{N}_{RL skill}\}$
\end{algorithmic}
\end{algorithm}

\begin{algorithm}[t]
\caption{Neuron-level Balance between Stability and Plasticity (NBSP) Applied in SAC}
\label{alg:algorithm_nbsp}
Initialize policy parameters $\theta$, Q-function parameters $\phi_1$, $\phi_2$, and target Q-function parameters $\phi_1'$, $\phi_2'$ \\
Initialize empty replay buffer $\mathcal{D}$ \\
Initialize replay interval $k$ \\
\textbf{\textcolor{myblue}{\textbf{Input:} Replay buffer $\mathbf{\mathcal{D}_{pre}}$, mask of the policy $\mathbf{mask_\theta}$  and mask of the Q-function parameters $\mathbf{mask_{\phi_1}, mask_{\phi_2}}$}}
\begin{algorithmic}[1] %[1] enables line numbers
\For{each task}
    \For{each iteration}
        \For{each environment step}
            \State Sample action $a_t \sim \pi_\theta(a_t|s_t)$
            \State Execute action $a_t$ and observe reward $r_t$ and next state $s_{t+1}$
            \State Store $(s_t, a_t, r_t, s_{t+1})$ in replay buffer $\mathcal{D}$
        \EndFor
        \For{each gradient step}
            \If{\textbf{\textcolor{myblue}{step $\mathbf{\equiv0\pmod{k}}$}}}{
                \textbf{\textcolor{myblue}{Sample batch of transitions $\mathbf{(s_i, a_i, r_i, s_{i+1})}$ from $\mathbf{\mathcal{D}_{pre}}$}} \;
            }
            \Else{
                Sample batch of transitions $(s_i, a_i, r_i, s_{i+1})$ from $\mathcal{D}$ \;
            }
            \EndIf
            
            \State Compute target value:
            \[
            y_i = r_i + \gamma \left( \min_{j=1,2} Q_{\phi_j'}(s_{i+1}, \tilde{a}_{i+1}) - \alpha \log \pi_\theta(\tilde{a}_{i+1}|s_{i+1}) \right)
            ,where \, \tilde{a}_{i+1} \sim \pi_\theta(\cdot|s_{i+1})
            \] 
            \State Update Q-functions by one step of gradient descent with mask:
            \[
            \phi_j \leftarrow \phi_j - \lambda_Q \mathbf{\textcolor{myblue}{mask_{\phi_j}}}\nabla_{\phi_j} \frac{1}{N} \sum_i \left( Q_{\phi_j}(s_i, a_i) - y_i \right)^2 \quad \text{for \,} j=1,2
            \]
            \State Update policy by one step of gradient ascent with mask:
            \[
            \theta \leftarrow \theta + \lambda_\pi \mathbf{\textcolor{myblue}{mask_\theta}}\nabla_\theta \frac{1}{N} \sum_i \left( \alpha \log \pi_\theta(a_i|s_i) - \min_{j=1,2} Q_{\phi_j}(s_i, a_i) \right)
            \]
            \State Update temperature $\alpha$ by one step of gradient descent:
            \[
            \alpha \leftarrow \alpha - \lambda_\alpha \nabla_\alpha \frac{1}{N} \sum_i \left( -\alpha \log \pi_\theta(a_i|s_i) - \alpha \bar{\mathcal{H}} \right)
            \]
            \State Update target Q-function parameters:
            \[
            \phi_j' \leftarrow \tau \phi_j + (1 - \tau) \phi_j' \quad \text{for } j=1,2
            \]
        \EndFor
    \EndFor
    \State \textbf{\textcolor{myblue}{Select RL skill neurons $\mathbf{\{\mathcal{N}_{RL \, skill}\}}$ according to Algorithm \ref{alg:algorithm_skill-neuron} }} 
    \State \textbf{\textcolor{myblue}{Update $\mathbf{mask_{\phi_1}}$,$\mathbf{mask_{\phi_2}}$ and $\mathbf{mask_\theta}$ using Eq. \ref{eq:mask}}}
    \State \textbf{\textcolor{myblue}{Store part of $\mathbf{\mathcal{D}}$ into $\mathbf{\mathcal{D}_{pre}}$ }}
\EndFor
\end{algorithmic}
\end{algorithm}

\section{Limitation and Future Work}
\textbf{Limitation}.
While the proposed NBSP method effectively balances stability and plasticity in DRL, it does have a notable limitation. Specifically, the number of RL skill neurons must be manually determined and adjusted according to the complexity of the learning task, as there is no automatic mechanism for this selection.

\textbf{Future work}. 
The neuron analysis introduced in this work offers a novel approach for identifying RL skill neurons, significantly enhancing the balance between stability and plasticity in DRL. The identification of RL skill neurons opens up several promising directions for future research and applications, such as: (1) Model Distillation: by focusing on RL skill neurons, it becomes possible to distill models by pruning  less relevant neurons, leading to more efficient and compact models with minimal performance degradation. (2) Bias Control and Model Manipulation: RL skill neurons could be leveraged to control biases and modify model behaviors by selectively adjusting their activations. This approach could be particularly valuable in scenarios requiring specific outputs or behaviors.

Regarding to the NBSP method, its applicable potential extends beyond DRL. It could also be adapted to other learning paradigms, such as supervised and unsupervised learning, to address similar stability-plasticity challenges.
In future work, we plan to explore these extensions and verify their effectiveness across various domains.

%%%%%%%%%%%%%%%%%%%%%%%%%%%%%%%%%%%%%%%%%%%%%%%%%%%%%%%%%%%%%%%%%%%%%%%%%%%%%%%
%%%%%%%%%%%%%%%%%%%%%%%%%%%%%%%%%%%%%%%%%%%%%%%%%%%%%%%%%%%%%%%%%%%%%%%%%%%%%%%

\end{document}